%% file: main.tex
\RequirePackage[OT1]{fontenc}
\documentclass[letterpaper, 10 pt, conference]{ieeeconf}  

\IEEEoverridecommandlockouts                              

\usepackage{graphics} 
\usepackage{amsmath} 
\usepackage{amssymb}  

\usepackage{caption}
\usepackage{subcaption}
\usepackage{booktabs}
\usepackage[style=ieee, url=false, doi=false, natbib=true, mincitenames=1, maxcitenames=1]{biblatex}

\usepackage{pgfplots}
\usepgfplotslibrary{groupplots,dateplot}
\usetikzlibrary{patterns,shapes.arrows, calc}
\pgfplotsset{compat=newest}

\pgfplotsset{every axis/.append style={
                     label style={font=\tiny},
                     tick label style={font=\small}  
                     }}
                     
\renewrobustcmd{\bfseries}{\fontseries{b}\selectfont}
\renewrobustcmd{\boldmath}{}
\newrobustcmd{\B}{\bfseries}

\usepackage{siunitx}
\usepackage{bm}

\newcommand{\R}{\mathbb{R}}

\DeclareMathAlphabet\mathbfcal{OMS}{cmsy}{b}{n}

\usepackage[breaklinks=true,colorlinks,bookmarks=false]{hyperref}
\usepackage[capitalise]{cleveref}

\usepackage{vector}
\usepackage{caption}
\title{\LARGE \bf
Attention Augmented ConvLSTM for Environment Prediction
}
\author{Bernard Lange, Masha Itkina, and Mykel J.~Kochenderfer$^{1}$
\thanks{*This work was supported by the Ford-Stanford Alliance.}
\thanks{$^{1}$Department of Aeronautics and Astronautics,
        Stanford University, Stanford, CA 94305, USA
        \texttt{\{blange, mitkina, mykel\}@stanford.edu}}%
}
\begin{document}

\maketitle
\thispagestyle{empty}
\pagestyle{empty}

\begin{abstract}
Safe and proactive planning in robotic systems generally requires accurate predictions of the environment. Prior work on environment prediction applied video frame prediction techniques to bird's-eye view environment representations, such as occupancy grids. ConvLSTM-based frameworks used previously often result in significant blurring of the predictions, loss of static environment structure, and vanishing of moving objects, thus hindering their applicability for use in safety-critical applications. In this work, we propose two extensions to the ConvLSTM architecture to address these issues. We present the Temporal Attention Augmented ConvLSTM (TAAConvLSTM) and Self-Attention Augmented ConvLSTM (SAAConvLSTM) frameworks for spatiotemporal occupancy grid prediction, and demonstrate improved performance over baseline architectures on the real-world KITTI and Waymo datasets. We provide our implementation at \url{ https://github.com/sisl/AttentionAugmentedConvLSTM}.
\end{abstract}

\input{sections/01-introduction} 
\input{sections/02-related_work}
\input{sections/03-background}
\input{sections/04-approach}

\input{sections/05-results}

\input{sections/06-discussion}

\section*{Acknowledgment} The authors would like to acknowledge this project being made possible by the funding from the Ford-Stanford Alliance. We thank Ransalu Senanayake for the valuable feedback and Mery Toyungyernsub for help with data processing.





{\small
\printbibliography
}
\clearpage
\input{sections/07-appendix}
\end{document}

%% file: sections/01-introduction.tex
\section{Introduction}
\label{sec:introduction}
Environment prediction is critical for developing intelligent agents capable of safe interaction with the surroundings, such as autonomous vehicles. Urban driving involves a multitude of complex scenarios that contain cluttered and obscured environments with a variety of moving and static obstacles. Experienced human drivers understand the semantics of the scene and can anticipate the future motion of other traffic participants. They implicitly use this knowledge to ensure the safety and comfort of their resulting maneuvers. Likewise, decision-making frameworks and trajectory planning algorithms must consider the inherent temporality of the perceived state and plan proactively to facilitate safe and comfortable transportation~\cite{fridovich2020confidence, itkina2019dynamic}.

Environment prediction architectures fall into two paradigms. A \textit{classical robotics approach} involves multiple sequential stages such as object detection and classification, state estimation, and motion modeling~\cite{petrovskaya2009model, ferguson2008detection, ferguson2013modifying, kuwata2008motion, dequaire2018deep}. However, this approach often requires full observability and operates on heavily preprocessed information. In contrast, a \textit{neural network-based approach} uses minimally processed information from sensors to directly perform environment prediction~\cite{chang1997environment, dequaire2018deep, hoermann2018dynamic}. 
We focus on the latter and exploit the rich perception data provided by LiDAR sensors. The task is posed as self-supervised sequence-to-sequence learning~\cite{sutskever2014sequence} in which a single scenario is decomposed into an input history sequence and a target prediction sequence. For the environment representation, we use a discrete map in the form of an occupancy grid~\cite{elfes1989using} due to its robustness to  partial environment observability~\cite{itkina2019dynamic}. This approach enables the use of uncertainty-aware occupancy state estimation, modeled using either Bayesian methods~\cite{elfes1989using} or Dempster-Shafer Theory (DST)~\cite{gordon1984dempster}, to update our belief about the surroundings.

\begin{figure}[t]
  \centering
  \centerline{\includegraphics[width=9cm]{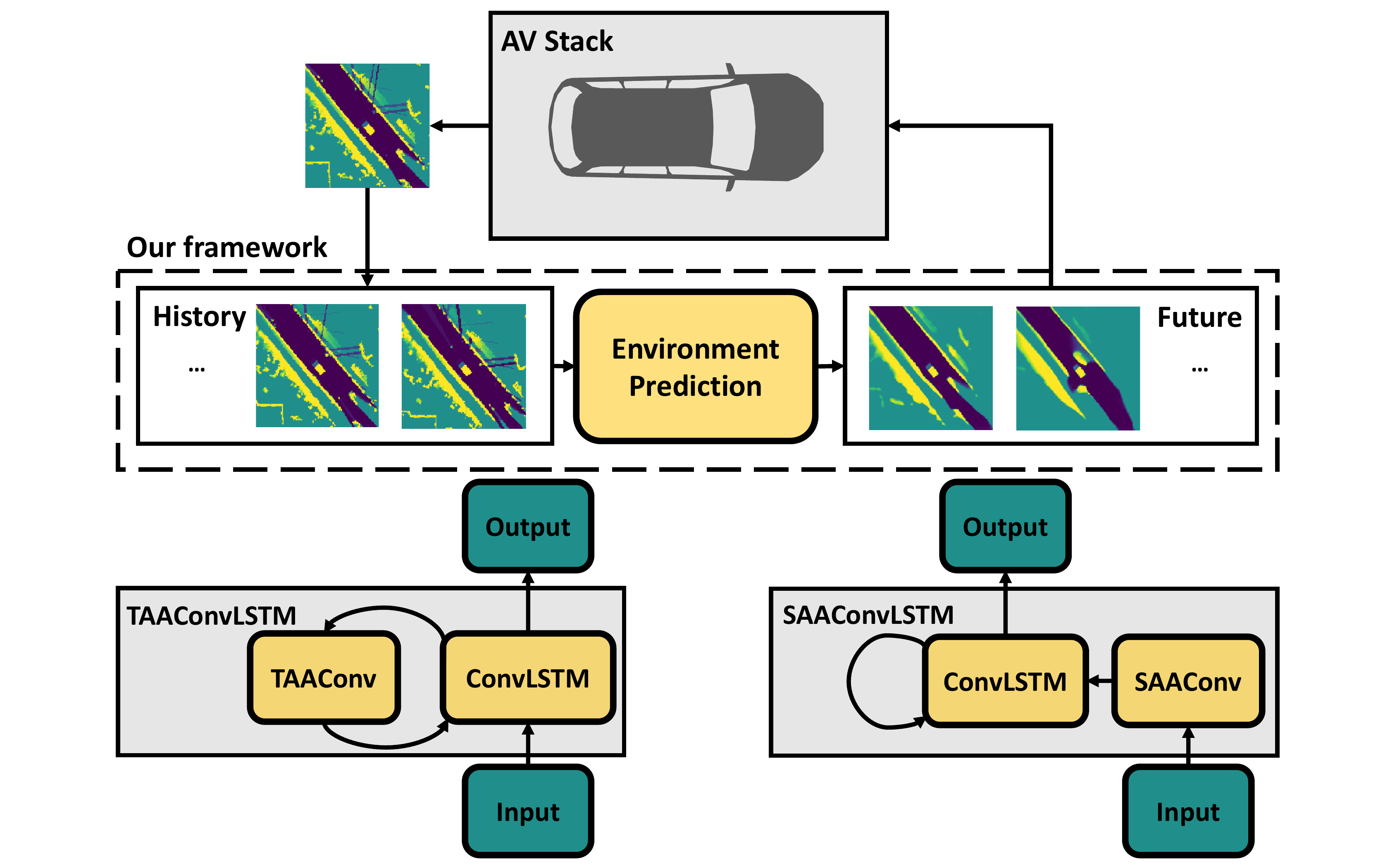}}
  \caption{We propose two novel attention-based extensions to the ConvLSTM architecture, TAAConvLSTM and SAAConvLSTM, for the sequence-to-sequence task of occupancy grid prediction in the autonomous driving setting.}
  \label{fig:Teaser}
  \vspace{-1.2em}
\end{figure}
Prior work used Convolutional Long Short Term Memory (ConvLSTM) frameworks~\cite{xingjian2015convolutional} to tackle occupancy grid prediction onboard both fixed~\cite{schreiber2019long} and moving~\cite{itkina2019dynamic, dequaire2018deep} platforms. These architectures are capable of capturing the dynamics of the environment, but they often suffer from mode-averaging and limited long-term dependencies between sequence elements~\cite{tang2019multiple}. Both issues lead to the blurriness of the predicted occupancy grid and, more importantly, the gradual disappearance of moving obstacles, which is a critical aspect of environment prediction~\cite{itkina2019dynamic, tang2019multiple}. 

Our method extends the approach for occupancy grid prediction defined by~\citet{itkina2019dynamic} with \textit{attention}~\cite{bahdanau2015neural, vaswani2017attention, parmar2019stand, bello2019attention} in order to improve the long-term dependencies captured by the prediction model. We propose attention-based extensions to the ConvLSTM architecture that aim to reduce prediction blurring and obstacle disappearance in longer time horizon predictions (see~\cref{fig:Teaser}). We hypothesize that prediction quality can be improved by strengthening the learned dependencies both across time and between different spatial regions in the environment. There are three subspaces to consider: feature (what is present in the environment), 2D spatial (where the object is in the environment), and temporal (when the object is present in the sequence). We introduce the Temporal Attention Augmented Convolution (TAAConv) operator, which
to our knowledge, is the first operator that successfully applies attention jointly to all three subspaces. 

We investigate two attention-based extensions of the ConvLSTM network where we replace convolution with our TAAConv operator and a self-attention equivalent (SAAConv). We compare the performance of the proposed architectures with other common spatiotemporal prediction models. The key contributions in this work are as follows:
\begin{itemize}
    \item We develop the novel Temporal Attention Augmented Convolution (TAAConv) operator that enables joint attention over the feature, 2D spatial, and temporal subspaces in a sequence of frames.
    \item We introduce the Temporal Attention Augmented ConvLSTM (TAAConvLSTM) and the Self-Attention Augmented ConvLSTM (SAAConvLSTM) architectures which encourage long-range dependencies between hidden representations and improve prediction quality for sequence-to-sequence environment prediction tasks.
    \item We apply the proposed mechanisms to local occupancy grid prediction for a moving ego-vehicle, and demonstrate superior performance compared to other state-of-the-art architectures, such as PredNet~\cite{lotter2016deep}, PredRNN++~\cite{wang2018predrnn++}, and CDNA~\cite{finn2016unsupervised}.
\end{itemize}
We evaluate the proposed mechanisms on two real-world datasets: KITTI~\cite{Geiger2013IJRR} and the Waymo Open Dataset~\cite{sun2020scalability}. We show that our architectures generate higher quality predictions by reducing blurring and moving obstacle disappearance, while better maintaining static environment structure in the predictions.

%% file: sections/02-related_work.tex
\section{Related Work}
\label{sec:related_work}
\textbf{Attention}: The \textit{attention} mechanism originated in the natural language processing field to create long-term dependencies between different parts of a sentence. \citet{bahdanau2015neural} showed that extending the conventional recurrent neural network (RNN) encoder-decoder architecture for transduction tasks with attention encourages long-term dependencies between different hidden representations at different sequence locations. \citet{vaswani2017attention} maintained the encoder-decoder approach but replaced the RNN model with a \textit{Transformer} consisting of a \textit{multi-head self-attention} mechanism. Several works have extended attention with convolution for the purposes of modeling sentence pairs~\cite{yin2016abcnn}, relation classification~\cite{wang2016relation}, and re-identification~\cite{xu2017jointly}. These approaches do not attend jointly to the feature, 2D spatial, and temporal subspaces, and, thus, are not directly applicable to spatiotemporal prediction tasks.

\citet{parmar2019stand} applied attention as a stand-alone vision operator by computing attention over local regions within an image. However, the lack of hardware-accelerated operations generally limits its application. \citet{bello2019attention} improved efficiency by reformulating self-attention to augment the convolution operation for visual tasks. We build on their approach for the task of occupancy grid prediction. We propose temporal attention and self-attention extensions to the ConvLSTM mechanism, whereby we redefine visual attention to apply it in the spatiotemporal setting, i.e. attention over the feature, 2D spatial, and temporal subspaces. Concurrently to our work, \citet{lin2020self} developed an architecture that introduces a self-attention-based memory component to the ConvLSTM, but does not enable direct temporal attention across different time steps as does our TAAConvLSTM architecture.

\textbf{Video Prediction}: \citet{srivastava2015unsupervised} proposed a sequence-to-sequence LSTM approach to video prediction.
\citet{xingjian2015convolutional} extended the LSTM model to capture spatial correlations by adding a convolution operator, thus forming the ubiquitous ConvLSTM. \citet{finn2016unsupervised} addressed the multimodality of the prediction task by considering an action-conditioned video prediction model termed CDNA. \citet{wang2018predrnn++} introduced the PredRNN++ architecture, which augments the ConvLSTM with \textit{zigzag} connections and increases recurrence depth to model complex short-term dynamics. \citet{lotter2016deep} proposed the Predictive Coding Network (PredNet), where the error signal between the prediction and the observation is propagated laterally and vertically within the recurrent architecture. These approaches focus on deterministic predictive models. A separate category of work tackles stochastic modelling (e.g. variational latent variable models~\cite{babaeizadeh2017stochastic}, adversarial training regimes~\cite{lee2018stochastic}, or normalizing flows~\cite{kumar2019videoflow}), which is beyond the scope of this paper. In this work, we apply deterministic video prediction techniques to occupancy grid prediction due to clear parallels between both tasks.

\textbf{Occupancy Grid Prediction}: Various adaptations of the RNN architecture with convolutions are commonly used for occupancy grid prediction tasks. \citet{dequaire2018deep} proposed a \textit{Deep Tracking} approach where an RNN with a Spatial Transformer was used for binary occupancy grid prediction. \citet{schreiber2019long} tested a ConvLSTM to predict occupancy grids for a stationary ego-vehicle in urban settings, such as intersections. \citet{mohajerin2019multi} employed a difference learning approach assuming a high refresh rate of the LiDAR sensor and high map similarity between frames. \citet{itkina2019dynamic} repurposed the PredNet architecture~\cite{lotter2016deep} for ego-centric occupancy grid prediction. This approach captured the relative dynamics of static and moving objects well but experienced significant blurring and the gradual disappearance of obstacles. In this paper, we formulate two attention-augmented ConvLSTM mechanisms and validate them on the best performing architecture for this task, PredNet~\cite{lotter2016deep, itkina2019dynamic}. Concurrent work by \citet{toyungyernsub2020double} addressed these issues with a double-prong architecture that assumes the static and dynamic objects in the scene to be known a priori. Unlike~\citet{toyungyernsub2020double}, we do not require access to highly accurate object detection and tracking capabilities.

%% file: sections/03-background.tex
\section{Background}
\label{sec:background}
\subsection{Convolutional Long Short-Term Memory (ConvLSTM)}
A \textit{ConvLSTM} is an RNN architecture that models the spatiotemporal correlations in a sequence. A single ConvLSTM unit for time step $t$ comprises the following elements: a \textit{memory cell} $\mathcal{C}_t$ accumulates information, an \textit{input gate} $i_t$ controls whether to include new information in the memory cell, a \textit{forget gate} $f_t$ maintains the flow of information from the previous time step, and an \textit{output gate} $o_t$ controls the information flow from the memory cell $\mathcal{C}_t$ to the \textit{hidden representation} $\mathcal{H}_t \in \mathbb{R}^{d \times H \times W}$. For an input $\mathcal{X}_t \in \mathbb{R}^{H \times W \times F_{in}}$, the ConvLSTM equations are denoted below, where `$*$' is the convolution operator and `$\circ$' is the Hadamard product~\cite{xingjian2015convolutional}: 
\begin{equation}
\small
    \label{eqn:convlstm}
    \begin{split}
        i_t &= \sigma(W_{xi} * \mathcal{X}_t + W_{hi} * \mathcal{H}_{t-1} + W_{ci} \circ \mathcal{C}_{t-1} + b_i) \\
        f_t &= \sigma(W_{xf} * \mathcal{X}_t + W_{hf} * \mathcal{H}_{t-1} + W_{cf} \circ \mathcal{C}_{t-1} + b_f) \\
        \mathcal{C}_t &= f_t \circ \mathcal{C}_{t-1} + i_t \circ \tanh(W_{xc} * \mathcal{X}_t + W_{hc} * \mathcal{H}_{t-1} + b_c) \\
        o_t &= \sigma(W_{xo} * \mathcal{X}_t + W_{ho} * \mathcal{H}_{t-1} + W_{co} \circ \mathcal{C}_t + b_o) \\
        \mathcal{H}_t &= o_t \circ \tanh(\mathcal{C}_t).
    \end{split}
\end{equation}
$W_{x(\cdot)}$, $W_{h(\cdot)} \in \mathbb{R}^{F_{in} \times k \times k \times d}$ and $W_{c(\cdot)}$, $b_{(\cdot)} \in \mathbb{R}^{d \times H \times W}$ denote the weights and biases of the architecture, where $k$ is the convolution kernel size.
\subsection{Self-Attention Augmented Convolution (SAAConv)}
\label{subsec:aaconv}
\citet{vaswani2017attention} introduced the self-attention mechanism to model sequential dependencies between elements in a position agnostic manner, which \citet{bello2019attention, parmar2019stand} later applied to visual tasks.
For a flattened image $\mathcal{X} \in \mathbb{R}^{HW \times F_{in}}$, they define a \textit{query}: $Q=\mathcal{X}W_q$, a \textit{key}: $K=\mathcal{X}W_k$, and a \textit{value}: $V=\mathcal{X}W_v$, where $W_q$, $W_k \in \mathbb{R}^{F_{in} \times d_k}$, and $W_v \in \mathbb{R}^{F_{in} \times d_v}$ are learned projection matrices for the input. Attention then selects or \textit{attends} to the subset of information encoded in the value $V$. The selection of information is determined by the softmax of the dot product between the query $Q$ and key $K$. As the query $Q$, key $K$, and value $V$ all derive from the original input $\mathcal{X}$, the mechanism dynamically regulates the flow of information that depends on the provided input $\mathcal{X}$ rather than solely the learned weights. Positional information can be added in the form of a learned positional encoding $S^{rel}_{H,W}$. Single-head attention is then evaluated as follows:
\begin{equation}
\small
    A_h(Q,K,V)_{ij} = \text{softmax} \bigg(\frac{Q_{i,:}K^T+(S^{rel}_{H,W})_{i,:}}{\sqrt{d^h_k}}\bigg)V_{:,j}.
\label{eqn:att}
\end{equation}
Multi-head attention concatenates the outputs $A_{h}$ for each head $h$ for a total of $N_h$ different heads, each with independent projection matrices $W^h_q$, $W^h_k$, and $W^h_v$, and fuses them using the learned matrix $W^o \in \mathbb{R}^{d_vN_h \times d_vN_h}$,
\begin{equation}
\small
    MA(Q,K,V) = [A_1, ..., A_{N_h}]W^{o}.
\label{eqn:mha}
\end{equation}
The motivation for multi-head attention is to enable attention in different representational subspaces which prevents the averaging effects present in single-head attention~\cite{vaswani2017attention}. The output can then be reshaped into a tensor of dimension $H \times W \times d_v$ and concatenated with the output of a convolution, forming Self-Attention Augmented Convolution (SAAConv)~\cite{bello2019attention}:
\begin{equation}
\small
 \mathcal{X} \oplus \mathcal{X} = [Conv(\mathcal{X}), MA(\mathcal{X}W_q, \mathcal{X}W_k, \mathcal{X}W_v)],
\label{eqn:aaconv}
\end{equation}
\noindent where $\mathcal{X} \oplus \mathcal{X}$ denotes the SAAConv operator on $\mathcal{X}$, which has a memory cost of ($O((HW)^2N_h)$)~\cite{bello2019attention}. 

%% file: sections/04-approach.tex
\section{Methodology}
\label{sec:approach}
\begin{figure}[htb]
  \centering
  \centerline{\includegraphics[width=9.5cm]{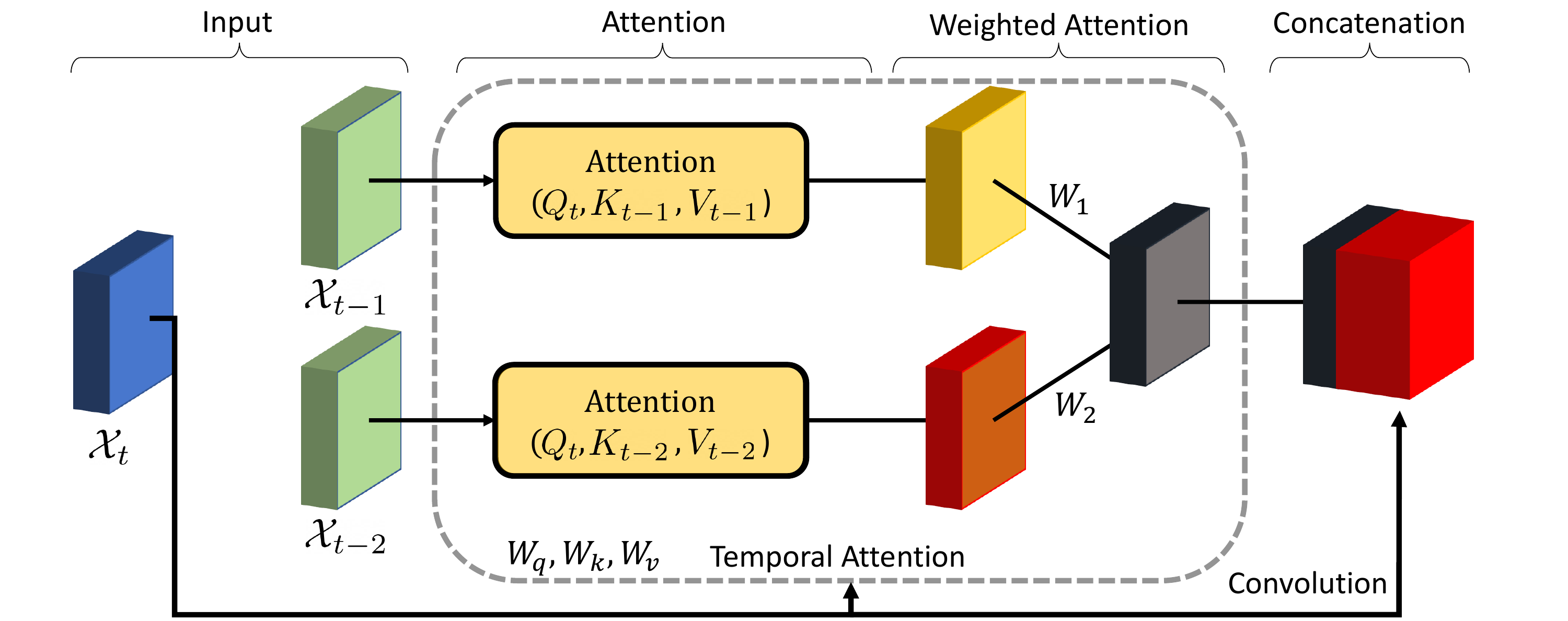}}
  \caption{The illustration shows TAAConv with a single head ($N_{h}=1$) and an attention horizon of two ($H_a = 2$).}
  \label{fig:AttConv}
  \vspace{-1.5em}
\end{figure}
Prior work on ConvLSTM-based frameworks has suffered from the gradual disappearance of moving agents within the predictions and blurring of the surrounding environment, making it unsuitable for longer time horizon predictions. 
We hypothesize that this phenomenon is caused by the incomplete hidden representation within the ConvLSTM architecture, which gradually loses the fine-grained details of dynamic objects and the static environment structure. We introduce the Temporal Attention Augmented Convolution (TAAConv) operator, which uses attention to draw visual dependencies between frames at different time steps. This operator is applied to the state-to-state transitions within the ConvLSTM, forming the Temporal Attention Augmented ConvLSTM (TAAConvLSTM). We posit that this mechanism reduces the loss of information that occurs in the hidden representation over time by recovering characteristics about the environment from previous, more accurate representations. Additionally, we propose the Self-Attention Augmented ConvLSTM (SAAConvLSTM), which applies the SAAConv operator to the input-to-state transitions of the ConvLSTM. We postulate that this self-attention extension provides a more accurate hidden representation than the vanilla ConvLSTM by dynamically regulating the input information depending on the scene content (i.e. highlighting important input features) which results in higher quality predictions.
\subsection{Problem Formulation}
We pose the environment prediction task as self-supervised sequence-to-sequence learning consisting of a history input sequence and a prediction output sequence. Given the history, our framework models how the environment evolves over time. For the environment representation, we use belief mass occupancy grids generated using DST~\cite{gordon1984dempster}, following the approach described by~\citet{itkina2019dynamic}. We define the belief mass occupancy grid $\mathcal{X}_t\in\R ^{2 \times H \times W}$, which contains the DST free and occupied belief mass values at time step $t$ for all grid cells. Let us denote the past $N$ frames as $\mathcal{X}_{t-N:t}$, which are sequentially provided to our architecture in order to build an internal, dynamic representation of the scene. After the $Nth$ frame, the model predicts the future $P$ frames, where $P$ is the prediction horizon. At each future time step $t+\tau$, the prediction $\mathcal{X}_{t+\tau}$ recursively serves as input at the next time step $t+\tau+1$ until time step $t+P$. Our objective is to predict $\mathcal{X}_{t+1:t+P}$.

\subsection{Temporal Attention Augmented Convolution (TAAConv)}
\label{subsec:taaconv}
We define the TAAConv operator to enable spatiotemporal attention over a sequence of frames (see~\cref{fig:AttConv}). The TAAConv draws direct dependencies between the current frame and the past $H_a$ frames, where $H_a$ is the attention horizon. We hypothesize this temporal attention module will learn to \textit{attend} to the essential moving and static objects in the frames across multiple time steps, thereby recovering missing details from earlier, less corrupted representations. Hence, the proposed TAAConv mechanism should improve the long-term dependencies captured by any recurrent prediction model.

Based on the current input $\mathcal{X}_t$ and the past input $\mathcal{X}_{t-\tau}$, a new query $Q_t$, key $K_{t-\tau}$, and value $V_{t-\tau}$ are computed following the attention for visual tasks approach outlined in \cref{subsec:aaconv}. Attention is then determined for each ($\mathcal{X}_t, \mathcal{X}_{t-\tau}$) pair, where $\tau \in 1:H_a$. Multi-head temporal attention is defined as follows:
\begin{equation}\label{eqn:spatiotemporaldattention}
\small
\begin{split}
MTA(\mathcal{X}_t,\mathcal{X}_{t-H_a:t-1}) &= \\ \sum_{\tau=t-H_a}^{t-1} w_{\tau} MA(\mathcal{X}_tW_q,&\mathcal{X}_{t-\tau}W_k,\mathcal{X}_{t-\tau}W_v),
\end{split}
\end{equation}
\noindent where $w_{\tau} \in \mathbb{R}$ are learned weights. The output is reshaped into a tensor of dimension $H \times W \times d_v$ and concatenated with the output of a convolutional layer. We define the TAAConv operator on $\mathcal{X}_t$ and $\mathcal{X}_{t-H_a:t-1}$ as follows:
\begin{equation}\label{eqn:taaconv}
\small
\begin{split}
 \mathcal{X}_{t} \odot \mathcal{X}_{t-H_a:t-1} = [Conv(\mathcal{X}_t), MTA(\mathcal{X}_t,\mathcal{X}_{t-H_a:t-1})],
\end{split}
\end{equation}
Hence, this method captures long range visual dependencies between the different time steps of a sequence by temporally weighting the attention scores based on the learned features and the spatial position, thereby attending to all three subspaces. The TAAConv operator has a memory cost of $O((HW)^2N_hH_a)$, which scales linearly in $H_a$ as compared to SAAConv.

\subsection{Temporal and Self-Attention Augmented ConvLSTM}
The prediction quality of a ConvLSTM network significantly deteriorates with longer time prediction horizons, resulting in substantial blurriness and moving object disappearance~\cite{lotter2016deep,wang2018predrnn++}. In this work, we formulate two extensions to the ConvLSTM mechanism to address these issues, \textit{TAAConvLSTM} and \textit{SAAConvLSTM}.

\textbf{TAAConvLSTM}: This approach is motivated by the assumption that the previous hidden representations $\mathcal{H}_{t-1-H_a:t-2}$ contain a more complete encoding of the moving object of interest than the most recent representation $\mathcal{H}_{t-1}$ due to compounding error. We use the TAAConv operator defined in \cref{subsec:taaconv} between $\mathcal{H}_{t-1}$ and $\mathcal{H}_{t-1-H_a:t-2}$ to recover detailed dynamic object predictions and environment structure, as a result, reducing predicted object vanishing and prediction blurring. We denote the time step range $t-1-H_a:t-2$ as $\Delta t$ for notational simplicity. The TAAConv operator (\textbf{bolded}) replaces the convolutional operators responsible for state-to-state transitions in the ConvLSTM (\cref{eqn:convlstm}):
\begin{equation} \label{eqn:tempattconvlstm}
\small
    \begin{split}
       i_t &= \sigma(W_{xi} * \mathcal{X}_t + \mathbfcal{H}_{t-1} \bm{\odot} \mathbfcal{H}_{\Delta t} + W_{ci} \circ \mathcal{C}_{t-1} + b_i) \\
        f_t &= \sigma(W_{xf} * \mathcal{X}_t +  \mathbfcal{H}_{t-1} \bm{\odot} \mathbfcal{H}_{\Delta t} + W_{cf} \circ \mathcal{C}_{t-1} + b_f) \\
        \mathcal{C}_t &= f_t \circ \mathcal{C}_{t-1} + i_t \circ \tanh(W_{xc} * \mathcal{X}_t + \bm{\mathbfcal{H}_{t-1} \odot \mathbfcal{H}_{\Delta t}} + b_c) \\
        o_t &= \sigma(W_{xo} * \mathcal{X}_t + \bm{\mathbfcal{H}_{t-1} \odot \mathbfcal{H}_{\Delta t}} + W_{co} \circ \mathcal{C}_t + b_o) \\
        \mathcal{H}_t &= o_t \circ \tanh(\mathcal{C}_t).
    \end{split}
\end{equation}

\textbf{SAAConvLSTM}: An alternative approach to mitigating object vanishing is to signalize which elements of the received $\mathcal{X}_t$ to preserve with the self-attention mechanism. We apply SAAConv (\cref{eqn:mha}) to the input-to-state transitions of the ConvLSTM (\cref{eqn:convlstm}). The self-attention mechanism allows for selective and dynamic regulation of incoming information to the hidden representation of the ConvLSTM, improving the flow of information by highlighting important input features. The SAAConvLSTM is formulated as follows (the SAAConv operator is \textbf{bolded}):
\begin{equation}
    \small
    \label{eqn:selfattconvlstm}
    \begin{split}
        i_t &= \sigma(\mathbfcal{X}_t \bm{\oplus} \mathbfcal{X}_t + W_{hi} * \mathcal{H}_{t-1} + W_{ci} \circ \mathcal{C}_{t-1} + b_i) \\
        f_t &= \sigma(\bm{\mathbfcal{X}_t \oplus \mathbfcal{X}_t} + W_{hf} * \mathcal{H}_{t-1} + W_{cf} \circ \mathcal{C}_{t-1} + b_f) \\
        \mathcal{C}_t &= f_t \circ \mathcal{C}_{t-1} + i_t \circ \tanh(\bm{\mathbfcal{X}_t \oplus \mathbfcal{X}_t} + W_{hc} * \mathcal{H}_{t-1} + b_c) \\
        o_t &= \sigma(\bm{\mathbfcal{X}_t \oplus \mathbfcal{X}_t} + W_{ho} * \mathcal{H}_{t-1} + W_{co} \circ \mathcal{C}_t + b_o) \\
        \mathcal{H}_t &= o_t \circ \tanh(\mathcal{C}_t).
    \end{split}
\end{equation}

%% file: sections/05-results.tex
\section{Experiments}
\label{sec:results}
We validate our proposed attention operators as part of the PredNet~\cite{lotter2016deep} architecture on real-world data.
The PredNet architecture was selected for direct comparison with prior environment prediction work~\cite{itkina2019dynamic}. We use the relative learned positional encoding defined by \citet{bello2019attention} in the attention calculation (\cref{eqn:att}). Our experiments demonstrate that the TAAConvLSTM and SAAConvLSTM mechanisms significantly reduce the vanishing of moving objects, improve static environment prediction, and outperform other state-of-the-art sequence-to-sequence prediction architectures, specifically PredNet~\cite{lotter2016deep}, PredRNN++~\cite{wang2018predrnn++}, and CDNA~\cite{finn2016unsupervised}.
The TAAConvLSTM is further examined in an ablation study where we investigate how each head of the temporal attention in the proposed TAAConv mechanism affects the predictions.
\input{figures/plot_KITTI}

\textbf{Experimental Setup}: We evaluate the capabilities of our proposed methods to make effective long-term predictions and compare them with other state-of-the-art sequence-to-sequence video prediction architectures on the real-world KITTI dataset~\cite{Geiger2013IJRR} and the Waymo Open Dataset~\cite{sun2020scalability}. Train-validation-test splits for the datasets are 0.9-0.02-0.08 and 0.75-0.10-0.15, respectively. In all experiments, we follow the procedure defined by \citet{itkina2019dynamic}, where we provide 5 past DST occupancy grids (\SI{0.5}{\second}) sequentially as input and make predictions for 15 future frames accounting for \SI{1.5}{\second} into the future. We consider occupancy grids of dimension $H \times W = 128\times 128$, corresponding to \SI{42.7}{\metre} $\times$ \SI{42.7}{\metre} with a \SI[parse-numbers=false]{0.\overline{3}}{\metre} resolution. All our models were trained using the PyTorch~\cite{paszke2019pytorch} library with the Adam~\cite{kingma2014adam} optimizer. Models were optimized for 200 epochs with 500 samples per epoch on an Nvidia RTX 2080Ti. We used the training procedure for multiple time step prediction defined by~\citet{lotter2016deep} that optimizes the $L1$ loss. We consider a four-layer PredNet architecture with TAAConvLSTM and SAAConvLSTM implemented in the top one and two layers, respectively. We selected the layers to place our mechanism from the top of the architecture, where the representation has the smallest size, until memory limitations were reached. All baselines were scaled in size to enable fair comparison between models.

\textbf{Evaluation Metrics}: We validate our prediction results using the traditional Mean Square Error (MSE) metric. However, evaluating the environment prediction task is challenging due to its multimodal output. Unobserved latent variables, such as intent, influence human decisions resulting in a multimodal future. Hence, we also consider the Image Similarity (IS) metric~\cite{birk2006merging}. IS calculates the smallest Manhattan distance between two grid cells with the same occupancy class assignment (free, occupied, and unknown). Thus, contrary to the MSE metric, which focuses on exact precision, IS successfully captures the variability in object positions between the ground truth and the predicted occupancy grids in the calculation of the score. Additionally, we construct a Moving Object Bounding Box Metric (MOBBM) to validate the reduced vanishing of moving objects over time. MOBBM evaluates the ratio of occupied cells present inside a vehicle's ground truth bounding box between the predicted and target frames for each time step. 
\subsection{KITTI Dataset}
\label{sec:kitti}
\begin{table}[t!]
    \caption{Comparison on the KITTI dataset (lower is better for all metrics). The top three models for each metric are \textbf{bolded}.}
      \centering
        \begin{tabular}{@{}lrrr@{}} \toprule
        \textbf{Model \& Hyperparameters} & \textbf{\#(M)} & \textbf{IS} & \textbf{MSE} ($\times 10^{-2}$) \\ \midrule
        TAAConvLSTM (ours) \\\midrule
        $H_a=2$, $N_h=4$ & 7.2 & 7.40 $\pm$ 0.07 & 3.74 $\pm$ 0.0013\\
        $H_a=6$, $N_h=4$ & 7.2 & \B 7.00 $ \pm$ \B 0.07 & 3.77 $\pm$ 0.0013  \\
        $H_a=4$, $N_h=2$ & 7.0 & 7.61 $\pm$ 0.07 & 3.92 $\pm$ 0.0013 \\
        $H_a=4$, $N_h=4$ & 7.2 & \B 6.91 $ \pm$ \B 0.06 & \B 3.62 $\pm $ \B 0.0012\\
        $H_a=4$, $N_h=6$ & 7.5 & 7.41 $\pm$ 0.07 & 3.82 $\pm$ 0.0013 \\
        $H_a=4$, $N_h=8$ & 7.7 & \B 7.02 $\pm$ \B 0.06 & 3.80 $\pm$ 0.0013\\ \midrule
        SAAConvLSTM (ours) \\ \midrule
        $N_h=2$ & 6.7 & 7.32 $\pm$ 0.06 & 3.67 $\pm$ 0.0012\\
        $N_h=4$ & 6.7 & 7.38 $\pm$ 0.07 & \B 3.57 $\pm$ \B 0.0012 \\
        $N_h=6$ & 6.7 & 7.32 $\pm$ 0.07 & 3.68 $\pm$ 0.0012 \\ \midrule
        PredNet \cite{lotter2016deep} & 6.9 & 7.68 $\pm$ 0.07 & 3.63 $\pm$ 0.0012 \\
        PredRNN++ \cite{wang2018predrnn++} & 7.2 & 8.86 $\pm$ 0.08 & \B 3.39 $\pm$ \B 0.0009 \\ 
        CDNA \cite{finn2016unsupervised} & 7.6 & 13.83 $\pm$ 0.14 & 4.12 $\pm$ 0.0011 \\ \midrule 
        \end{tabular}
    \label{table:Kittitab}
    \vspace{-2em}
\end{table}
We compare the prediction performance of our proposed TAAConvLSTM and SAAConvLSTM architectures with the considered baselines on the KITTI dataset~\cite{Geiger2013IJRR} in \cref{table:Kittitab}. We vary the attention horizon $H_a$ and the number of heads $N_h$ in TAAConvLSTM and the number of heads $N_h$ in SAAConvLSTM. An increase in the attention horizon and the number of heads within TAAConvLSTM improves quantitative performance until reaching diminishing returns at $H_a=4$ and $N_h=4$. The performance of SAAConvLSTM is less sensitive to the number of attention heads selected. All our proposed models outperform the baselines in terms of the IS metric by a significant margin. However, PredRNN++ outperforms in the MSE metric due to its blurring of the predicted occupancy grids. The trade-off between IS and MSE metrics is further discussed in \cref{sec:discussion}. 
\begin{figure}[t!]
     \centering
     \input{figures/IS_MOBBM.tex}
     \caption{\small MOBBM (top, higher is better) and IS (bottom, lower is better) metric values versus the prediction horizon for a maximum of \SI{2.5}{\second} ahead. The proposed TAAConvLSTM and SAAConvLSTM architectures outperform all baselines at longer prediction horizons. The performance gap in IS between TAAConvLSTM and the baselines increases with longer time horizon predictions. We note that the CDNA architecture is not included in the IS plot as its performance is significantly worse than the other models.}
     \label{fig:IS_MOBBM}
     \vspace{-2em}
\end{figure}

\cref{fig:KittiVis} shows qualitative prediction performance for \SI{2.5}{\second} ahead. By considering a longer time prediction horizon than the \SI{1.5}{\second} horizon used during training, we evaluate the generalization capabilities of the models. \cref{fig:KittiVis} depicts two example urban scenarios: the ego-vehicle passing an oncoming vehicle in a two lane road and a turn at an intersection. Our proposed architectures visually demonstrate a significant reduction in the vanishing of moving objects and result in more accurate occlusion propagation in the predictions. An overall increase in the quality of the predictions is observed. TAAConvLSTM excels at the motion propagation of moving objects (\cref{fig:KittiVis} Left). In addition to reducing the vanishing of moving objects in the predictions, SAAConvLSTM models the ego-vehicle interaction with the environment more accurately, e.g. the ego-vehicle position in the intersection while turning (\cref{fig:KittiVis} Right). All of the baselines failed to correctly propagate the motion of the moving objects and maintain the structure of the intersection. 

To evaluate the improvements in moving object retention in the predictions and the long-term dependency representation, we plot the MOBBM and IS metrics over time in \cref{fig:IS_MOBBM}. Both our models achieve better MOBBM and IS metric values than the baselines, indicating more accurate moving object prediction and reduced object disappearance, particularly at longer time horizon predictions. 

\subsection{Waymo Open Dataset}
\label{sec:waymo}
\input{figures/plot_Waymo}
We further validate our best performing models in \cref{table:Kittitab} on the much larger Waymo dataset~\cite{sun2020scalability} and compare their performance to the baseline architectures. \cref{fig:WaymoVis} depicts two example scenarios with interactions at or close to an intersection. In both cases, TAAConvLSTM and SAAConvLSTM visually outperform all baselines by more accurately predicting the static environment transformation, correctly predicting the propagation of occlusions, and maintaining the presence of the moving objects in the scene. We compare the models quantitatively in \cref{tab:Waymo}. We observe that SAAConvLSTM underperforms numerically even though qualitatively it performs well. TAAConvLSTM has a superior IS score but is outperformed by PredRNN++ and CDNA on the MSE metric due to the significant blurriness of their predictions, as further discussed in \cref{sec:discussion}.

\begin{table}[h]
\centering
\caption{Comparison on the Waymo Open Dataset (lower is better for all metrics). The best performing models are \textbf{bolded}.}
\begin{tabular}{@{}lrrr@{}} \toprule
\textbf{Model} & \textbf{\#(M)} & \textbf{IS} & \textbf{MSE} ($\times 10^{-3}$)\\ \midrule
TAAConvLSTM (ours) & 7.2  & \B 3.51 $\pm$ \B 0.04  & $2.33 \pm 0.0063$ \\
SAAConvLSTM (ours) & 6.7 & $3.64 \pm 0.04$  & $2.39 \pm 0.0095$  \\ \midrule
PredNet \cite{lotter2016deep} & 6.9 & $3.56 \pm 0.04$ & $2.34 \pm 0.0063$  \\
PredRNN++ \cite{wang2018predrnn++} & 7.2 & $4.68 \pm  0.05$ & \B 2.25 $\pm$ \B 0.0052  \\
CDNA \cite{finn2016unsupervised} & 7.6 & $5.49 \pm  0.05$ & $2.31 \pm 0.0051$  \\
\bottomrule
\end{tabular}
\label{tab:Waymo}
\vspace{-1.5em}
\end{table}
\subsection{Ablation Study}
\begin{figure}[h]
  \centering
  \centerline{\includegraphics[width=7.3cm]{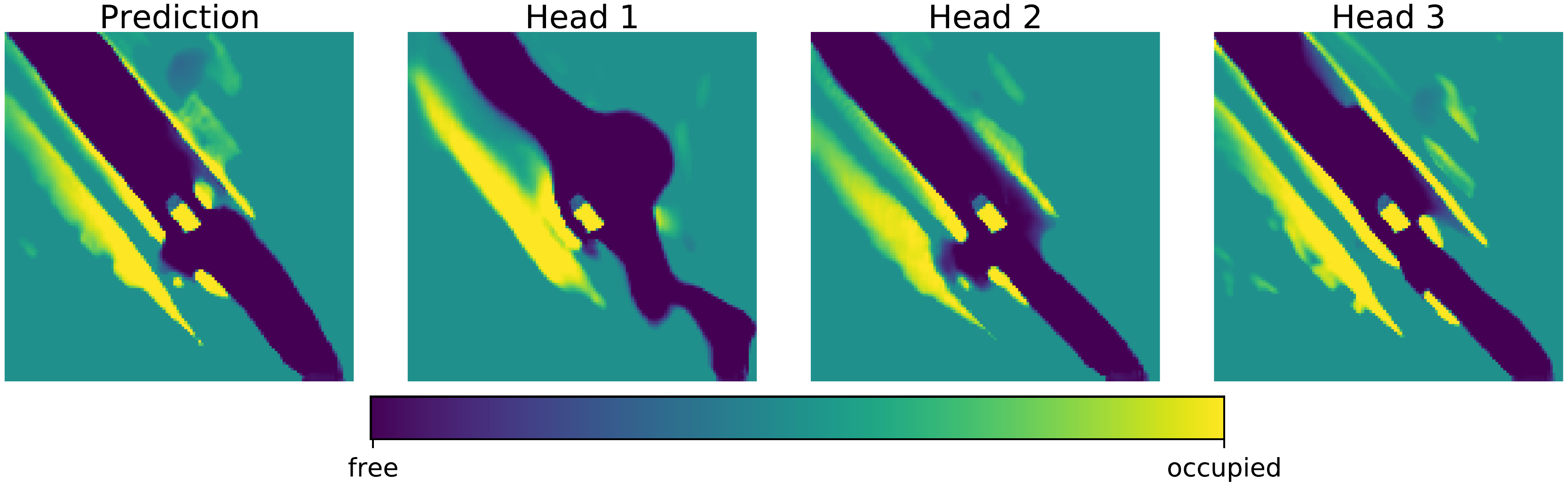}}
  \caption{We conduct an ablation study to better understand the proposed TAAConv mechanism as part of the TAAConvLSTM architecture. From left to right: the full prediction, prediction with the first, second, and third attention heads removed.}
  \label{fig:Ablation}
  \vspace{-1.5em}
\end{figure}

We validate our proposed TAAConv mechanism and investigate how each attention head affects the generated predictions as part of the proposed TAAConvLSTM architecture. We use a smaller variant of our architecture, 3-layer PredNet with TAAConvLSTM at the top layer with $N_h=3$ and $H_a=4$, to better understand the workings of our proposed approach. To evaluate how each head contributes to the prediction, we zero out the output of each head one at a time in the attention mechanism and perform the prediction as usual. As shown in \cref{fig:Ablation}, multi-head temporal attention distinctly distributes tasks among the attention heads. Removal of each of the three heads results in: 1) loss of the static environment structure, 2) loss of moving objects, 3) translation offset of the moving objects resulting in a delayed prediction. We hypothesize that the third head learned the motion of the moving objects. Thus, the conducted ablation study illustrates the multimodal capabilities of the attention mechanism, and demonstrates that the proposed mechanism plays an important role in our modified PredNet architecture. 

%% file: figures/plot_KITTI.tex
\begin{figure*}[h]
\begin{center}
\scalebox{0.85}{
\begin{tikzpicture} [font=\tiny]
\begin{groupplot}[
   group style={
      group name=left plots,
      group size=5 by 1,
      horizontal sep=0pt,
      x descriptions at=edge bottom},
   width=1.25cm,
   height=1.25cm,
   scale only axis,
   every axis title/.style={yshift=4pt, xshift=17.7pt}]

\nextgroupplot[
title={$t=-4$},
ylabel={\scalebox{.8}{Input}},
axis background/.style={fill=white!89.8039215686275!black},
axis line style={black},
xmajorgrids,
xmajorticks=false,
xmin=-0.5, xmax=127.5,
y dir=reverse,
ymajorgrids,
ymajorticks=false,
ymin=-0.5, ymax=127.5
]
\addplot graphics [includegraphics cmd=\pgfimage,xmin=-0.5, xmax=127.5, ymin=127.5, ymax=-0.5] {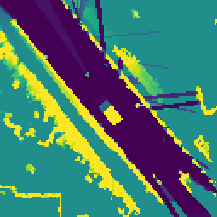};   
   
\nextgroupplot[
title={$t=-3$},
axis background/.style={fill=white!89.8039215686275!black},
axis line style={black},
xmajorgrids,
xmajorticks=false,
xmin=-0.5, xmax=127.5,
y dir=reverse,
ymajorgrids,
ymajorticks=false,
ymin=-0.5, ymax=127.5
]
\addplot graphics [includegraphics cmd=\pgfimage,xmin=-0.5, xmax=127.5, ymin=127.5, ymax=-0.5] {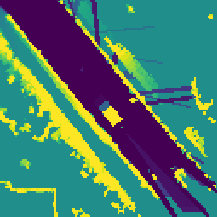};

\nextgroupplot[
title={$t=-2$},
axis background/.style={fill=white!89.8039215686275!black},
axis line style={black},
xmajorgrids,
xmajorticks=false,
xmin=-0.5, xmax=127.5,
y dir=reverse,
ymajorgrids,
ymajorticks=false,
ymin=-0.5, ymax=127.5
]
\addplot graphics [includegraphics cmd=\pgfimage,xmin=-0.5, xmax=127.5, ymin=127.5, ymax=-0.5] {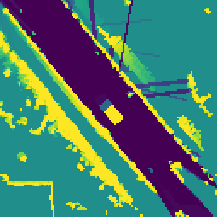};

\nextgroupplot[
title={$t=-1$},
axis background/.style={fill=white!89.8039215686275!black},
axis line style={black},
xmajorgrids,
xmajorticks=false,
xmin=-0.5, xmax=127.5,
y dir=reverse,
ymajorgrids,
ymajorticks=false,
ymin=-0.5, ymax=127.5
]
\addplot graphics [includegraphics cmd=\pgfimage,xmin=-0.5, xmax=127.5, ymin=127.5, ymax=-0.5] {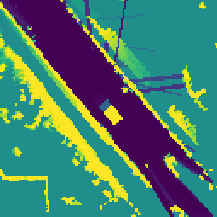};

\nextgroupplot[
title={$t=0$},
axis background/.style={fill=white!89.8039215686275!black},
axis line style={black},
xmajorgrids,
xmajorticks=false,
xmin=-0.5, xmax=127.5,
y dir=reverse,
ymajorgrids,
ymajorticks=false,
ymin=-0.5, ymax=127.5
]
\addplot graphics [includegraphics cmd=\pgfimage,xmin=-0.5, xmax=127.5, ymin=127.5, ymax=-0.5] {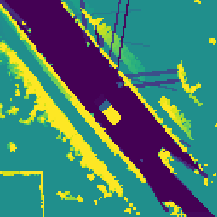};
\end{groupplot}

\begin{groupplot}[
    group style={
        group name=left bottom,
       group size= 5 by 6,
       horizontal sep=0pt,
       vertical sep = 0pt,
       x descriptions at=edge bottom},
    width=1.25cm,
    height=1.25cm,
    scale only axis,
    ytick pos=left,
    every axis title/.style={yshift=4pt, xshift=17.7pt}]
 \nextgroupplot[
    anchor=north west, at={($(left plots c1r1.south west) - (0.0cm,0.6cm)$)},
    title={$t=5$},
    ylabel={\scalebox{.8}{Ground truth}},
    axis background/.style={fill=white!89.8039215686275!black},
    axis line style={black},
    xmajorgrids,
    xmajorticks=false,
    xmin=-0.5, xmax=127.5,
    y dir=reverse,
    ymajorgrids,
    ymajorticks=false,
    ymin=-0.5, ymax=127.5
    ]
 \addplot graphics [includegraphics cmd=\pgfimage,xmin=-0.5, xmax=127.5, ymin=127.5, ymax=-0.5] {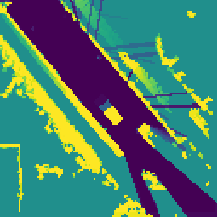};
 
  \nextgroupplot[
    title={$t=10$},
    axis background/.style={fill=white!89.8039215686275!black},
    axis line style={black},
    xmajorgrids,
    xmajorticks=false,
    xmin=-0.5, xmax=127.5,
    y dir=reverse,
    ymajorgrids,
    ymajorticks=false,
    ymin=-0.5, ymax=127.5
    ]
 \addplot graphics [includegraphics cmd=\pgfimage,xmin=-0.5, xmax=127.5, ymin=127.5, ymax=-0.5] {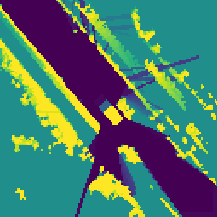};
 
 \nextgroupplot[
    title={$t=15$},
    axis background/.style={fill=white!89.8039215686275!black},
    axis line style={black},
    xmajorgrids,
    xmajorticks=false,
    xmin=-0.5, xmax=127.5,
    y dir=reverse,
    ymajorgrids,
    ymajorticks=false,
    ymin=-0.5, ymax=127.5
    ]
 \addplot graphics [includegraphics cmd=\pgfimage,xmin=-0.5, xmax=127.5, ymin=127.5, ymax=-0.5] {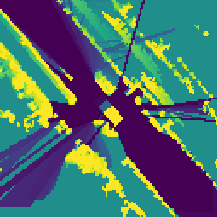};
 \nextgroupplot[
    title={$t=20$},
    axis background/.style={fill=white!89.8039215686275!black},
    axis line style={black},
    xmajorgrids,
    xmajorticks=false,
    xmin=-0.5, xmax=127.5,
    y dir=reverse,
    ymajorgrids,
    ymajorticks=false,
    ymin=-0.5, ymax=127.5
    ]
 \addplot graphics [includegraphics cmd=\pgfimage,xmin=-0.5, xmax=127.5, ymin=127.5, ymax=-0.5] {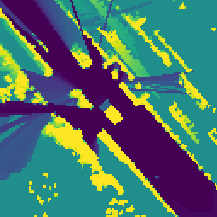};
 \nextgroupplot[
    title={$t=25$},
    axis background/.style={fill=white!89.8039215686275!black},
    axis line style={black},
    xmajorgrids,
    xmajorticks=false,
    xmin=-0.5, xmax=127.5,
    y dir=reverse,
    ymajorgrids,
    ymajorticks=false,
    ymin=-0.5, ymax=127.5
    ]
 \addplot graphics [includegraphics cmd=\pgfimage,xmin=-0.5, xmax=127.5, ymin=127.5, ymax=-0.5] {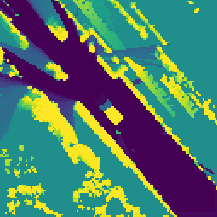};
\nextgroupplot[
    ylabel={\scalebox{.8}{TAAConvLSTM}},
    axis background/.style={fill=white!89.8039215686275!black},
    axis line style={black},
    xmajorgrids,
    xmajorticks=false,
    xmin=-0.5, xmax=127.5,
    y dir=reverse,
    ymajorgrids,
    ymajorticks=false,
    ymin=-0.5, ymax=127.5
    ]
 \addplot graphics [includegraphics cmd=\pgfimage,xmin=-0.5, xmax=127.5, ymin=127.5, ymax=-0.5] {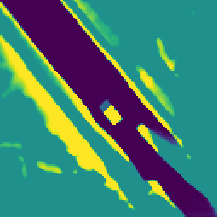};
 \nextgroupplot[
    axis background/.style={fill=white!89.8039215686275!black},
    axis line style={black},
    xmajorgrids,
    xmajorticks=false,
    xmin=-0.5, xmax=127.5,
    y dir=reverse,
    ymajorgrids,
    ymajorticks=false,
    ymin=-0.5, ymax=127.5
    ]
 \addplot graphics [includegraphics cmd=\pgfimage,xmin=-0.5, xmax=127.5, ymin=127.5, ymax=-0.5] {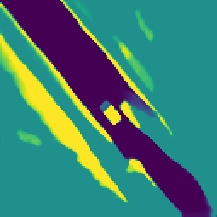};
 \nextgroupplot[
    axis background/.style={fill=white!89.8039215686275!black},
    axis line style={black},
    xmajorgrids,
    xmajorticks=false,
    xmin=-0.5, xmax=127.5,
    y dir=reverse,
    ymajorgrids,
    ymajorticks=false,
    ymin=-0.5, ymax=127.5
    ]
 \addplot graphics [includegraphics cmd=\pgfimage,xmin=-0.5, xmax=127.5, ymin=127.5, ymax=-0.5] {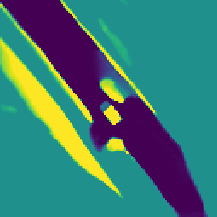};
 \nextgroupplot[
    axis background/.style={fill=white!89.8039215686275!black},
    axis line style={black},
    xmajorgrids,
    xmajorticks=false,
    xmin=-0.5, xmax=127.5,
    y dir=reverse,
    ymajorgrids,
    ymajorticks=false,
    ymin=-0.5, ymax=127.5
    ]
 \addplot graphics [includegraphics cmd=\pgfimage,xmin=-0.5, xmax=127.5, ymin=127.5, ymax=-0.5] {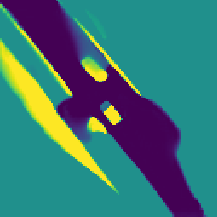};
 \nextgroupplot[
    axis background/.style={fill=white!89.8039215686275!black},
    axis line style={black},
    xmajorgrids,
    xmajorticks=false,
    xmin=-0.5, xmax=127.5,
    y dir=reverse,
    ymajorgrids,
    ymajorticks=false,
    ymin=-0.5, ymax=127.5
    ]
 \addplot graphics [includegraphics cmd=\pgfimage,xmin=-0.5, xmax=127.5, ymin=127.5, ymax=-0.5] {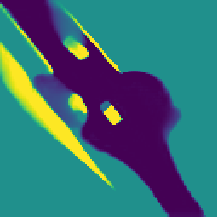};
\nextgroupplot[
    ylabel={\scalebox{.8}{SAAConvLSTM}},
    axis background/.style={fill=white!89.8039215686275!black},
    axis line style={black},
    xmajorgrids,
    xmajorticks=false,
    xmin=-0.5, xmax=127.5,
    y dir=reverse,
    ymajorgrids,
    ymajorticks=false,
    ymin=-0.5, ymax=127.5
    ]
 \addplot graphics [includegraphics cmd=\pgfimage,xmin=-0.5, xmax=127.5, ymin=127.5, ymax=-0.5] {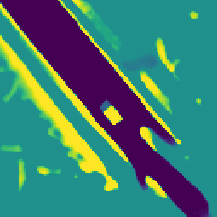};
 \nextgroupplot[
    axis background/.style={fill=white!89.8039215686275!black},
    axis line style={black},
    xmajorgrids,
    xmajorticks=false,
    xmin=-0.5, xmax=127.5,
    y dir=reverse,
    ymajorgrids,
    ymajorticks=false,
    ymin=-0.5, ymax=127.5
    ]
 \addplot graphics [includegraphics cmd=\pgfimage,xmin=-0.5, xmax=127.5, ymin=127.5, ymax=-0.5] {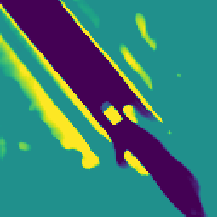};
 \nextgroupplot[
    axis background/.style={fill=white!89.8039215686275!black},
    axis line style={black},
    xmajorgrids,
    xmajorticks=false,
    xmin=-0.5, xmax=127.5,
    y dir=reverse,
    ymajorgrids,
    ymajorticks=false,
    ymin=-0.5, ymax=127.5
    ]
 \addplot graphics [includegraphics cmd=\pgfimage,xmin=-0.5, xmax=127.5, ymin=127.5, ymax=-0.5] {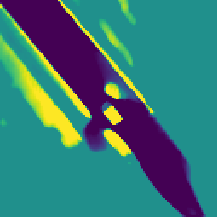};
 \nextgroupplot[
    axis background/.style={fill=white!89.8039215686275!black},
    axis line style={black},
    xmajorgrids,
    xmajorticks=false,
    xmin=-0.5, xmax=127.5,
    y dir=reverse,
    ymajorgrids,
    ymajorticks=false,
    ymin=-0.5, ymax=127.5
    ]
 \addplot graphics [includegraphics cmd=\pgfimage,xmin=-0.5, xmax=127.5, ymin=127.5, ymax=-0.5] {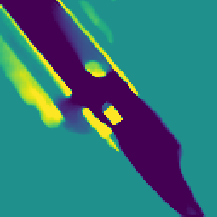};
 \nextgroupplot[
    axis background/.style={fill=white!89.8039215686275!black},
    axis line style={black},
    xmajorgrids,
    xmajorticks=false,
    xmin=-0.5, xmax=127.5,
    y dir=reverse,
    ymajorgrids,
    ymajorticks=false,
    ymin=-0.5, ymax=127.5
    ]
 \addplot graphics [includegraphics cmd=\pgfimage,xmin=-0.5, xmax=127.5, ymin=127.5, ymax=-0.5] {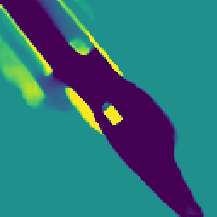};
\nextgroupplot[
    ylabel={\scalebox{.8}{PredNet}},
    axis background/.style={fill=white!89.8039215686275!black},
    axis line style={black},
    xmajorgrids,
    xmajorticks=false,
    xmin=-0.5, xmax=127.5,
    y dir=reverse,
    ymajorgrids,
    ymajorticks=false,
    ymin=-0.5, ymax=127.5
    ]
 \addplot graphics [includegraphics cmd=\pgfimage,xmin=-0.5, xmax=127.5, ymin=127.5, ymax=-0.5] {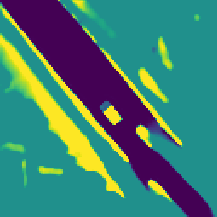};
 \nextgroupplot[
    axis background/.style={fill=white!89.8039215686275!black},
    axis line style={black},
    xmajorgrids,
    xmajorticks=false,
    xmin=-0.5, xmax=127.5,
    y dir=reverse,
    ymajorgrids,
    ymajorticks=false,
    ymin=-0.5, ymax=127.5
    ]
 \addplot graphics [includegraphics cmd=\pgfimage,xmin=-0.5, xmax=127.5, ymin=127.5, ymax=-0.5] {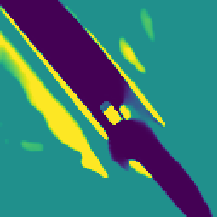};
 \nextgroupplot[
    axis background/.style={fill=white!89.8039215686275!black},
    axis line style={black},
    xmajorgrids,
    xmajorticks=false,
    xmin=-0.5, xmax=127.5,
    y dir=reverse,
    ymajorgrids,
    ymajorticks=false,
    ymin=-0.5, ymax=127.5
    ]
 \addplot graphics [includegraphics cmd=\pgfimage,xmin=-0.5, xmax=127.5, ymin=127.5, ymax=-0.5] {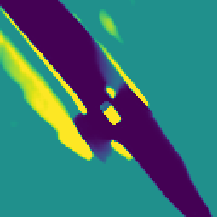};
 \nextgroupplot[
    axis background/.style={fill=white!89.8039215686275!black},
    axis line style={black},
    xmajorgrids,
    xmajorticks=false,
    xmin=-0.5, xmax=127.5,
    y dir=reverse,
    ymajorgrids,
    ymajorticks=false,
    ymin=-0.5, ymax=127.5
    ]
 \addplot graphics [includegraphics cmd=\pgfimage,xmin=-0.5, xmax=127.5, ymin=127.5, ymax=-0.5] {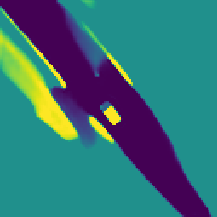};
 \nextgroupplot[
    axis background/.style={fill=white!89.8039215686275!black},
    axis line style={black},
    xmajorgrids,
    xmajorticks=false,
    xmin=-0.5, xmax=127.5,
    y dir=reverse,
    ymajorgrids,
    ymajorticks=false,
    ymin=-0.5, ymax=127.5
    ]
 \addplot graphics [includegraphics cmd=\pgfimage,xmin=-0.5, xmax=127.5, ymin=127.5, ymax=-0.5] {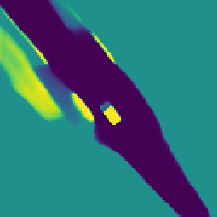};
\nextgroupplot[
    ylabel={\scalebox{.8}{PredRNN++}},
    axis background/.style={fill=white!89.8039215686275!black},
    axis line style={black},
    xmajorgrids,
    xmajorticks=false,
    xmin=-0.5, xmax=127.5,
    y dir=reverse,
    ymajorgrids,
    ymajorticks=false,
    ymin=-0.5, ymax=127.5
    ]
 \addplot graphics [includegraphics cmd=\pgfimage,xmin=-0.5, xmax=127.5, ymin=127.5, ymax=-0.5] {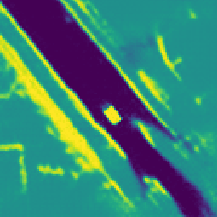};
 
 \nextgroupplot[
    axis background/.style={fill=white!89.8039215686275!black},
    axis line style={black},
    xmajorgrids,
    xmajorticks=false,
    xmin=-0.5, xmax=127.5,
    y dir=reverse,
    ymajorgrids,
    ymajorticks=false,
    ymin=-0.5, ymax=127.5
    ]
 \addplot graphics [includegraphics cmd=\pgfimage,xmin=-0.5, xmax=127.5, ymin=127.5, ymax=-0.5] {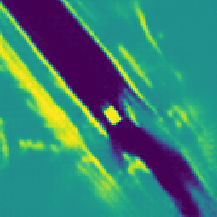};
 
 \nextgroupplot[
    axis background/.style={fill=white!89.8039215686275!black},
    axis line style={black},
    xmajorgrids,
    xmajorticks=false,
    xmin=-0.5, xmax=127.5,
    y dir=reverse,
    ymajorgrids,
    ymajorticks=false,
    ymin=-0.5, ymax=127.5
    ]
 \addplot graphics [includegraphics cmd=\pgfimage,xmin=-0.5, xmax=127.5, ymin=127.5, ymax=-0.5] {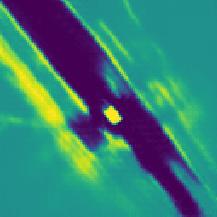};
 
 \nextgroupplot[
    axis background/.style={fill=white!89.8039215686275!black},
    axis line style={black},
    xmajorgrids,
    xmajorticks=false,
    xmin=-0.5, xmax=127.5,
    y dir=reverse,
    ymajorgrids,
    ymajorticks=false,
    ymin=-0.5, ymax=127.5
    ]
 \addplot graphics [includegraphics cmd=\pgfimage,xmin=-0.5, xmax=127.5, ymin=127.5, ymax=-0.5] {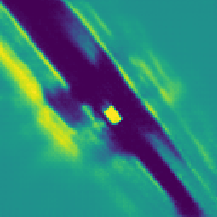};
 
 \nextgroupplot[
    axis background/.style={fill=white!89.8039215686275!black},
    axis line style={black},
    xmajorgrids,
    xmajorticks=false,
    xmin=-0.5, xmax=127.5,
    y dir=reverse,
    ymajorgrids,
    ymajorticks=false,
    ymin=-0.5, ymax=127.5
    ]
 \addplot graphics [includegraphics cmd=\pgfimage,xmin=-0.5, xmax=127.5, ymin=127.5, ymax=-0.5] {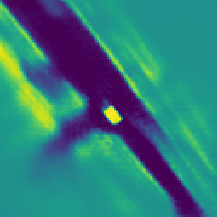};
 
  \nextgroupplot[
    axis background/.style={fill=white!89.8039215686275!black},
    axis line style={black},
    ylabel={\scalebox{.8}{CDNA}},
    xmajorgrids,
    xmajorticks=false,
    xmin=-0.5, xmax=127.5,
    y dir=reverse,
    ymajorgrids,
    ymajorticks=false,
    ymin=-0.5, ymax=127.5
    ]
 \addplot graphics [includegraphics cmd=\pgfimage,xmin=-0.5, xmax=127.5, ymin=127.5, ymax=-0.5] {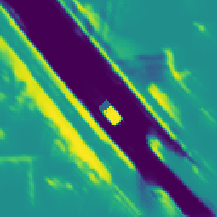};
 
  \nextgroupplot[
    axis background/.style={fill=white!89.8039215686275!black},
    axis line style={black},
    xmajorgrids,
    xmajorticks=false,
    xmin=-0.5, xmax=127.5,
    y dir=reverse,
    ymajorgrids,
    ymajorticks=false,
    ymin=-0.5, ymax=127.5
    ]
 \addplot graphics [includegraphics cmd=\pgfimage,xmin=-0.5, xmax=127.5, ymin=127.5, ymax=-0.5] {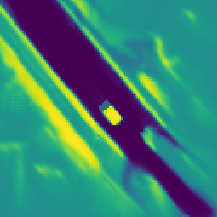};
 
  \nextgroupplot[
    axis background/.style={fill=white!89.8039215686275!black},
    axis line style={black},
    xmajorgrids,
    xmajorticks=false,
    xmin=-0.5, xmax=127.5,
    y dir=reverse,
    ymajorgrids,
    ymajorticks=false,
    ymin=-0.5, ymax=127.5
    ]
 \addplot graphics [includegraphics cmd=\pgfimage,xmin=-0.5, xmax=127.5, ymin=127.5, ymax=-0.5] {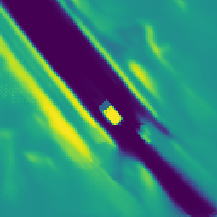};
 
  \nextgroupplot[
    axis background/.style={fill=white!89.8039215686275!black},
    axis line style={black},
    xmajorgrids,
    xmajorticks=false,
    xmin=-0.5, xmax=127.5,
    y dir=reverse,
    ymajorgrids,
    ymajorticks=false,
    ymin=-0.5, ymax=127.5
    ]
 \addplot graphics [includegraphics cmd=\pgfimage,xmin=-0.5, xmax=127.5, ymin=127.5, ymax=-0.5] {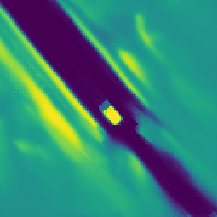};
 
  \nextgroupplot[
    axis background/.style={fill=white!89.8039215686275!black},
    axis line style={black},
    xmajorgrids,
    xmajorticks=false,
    xmin=-0.5, xmax=127.5,
    y dir=reverse,
    ymajorgrids,
    ymajorticks=false,
    ymin=-0.5, ymax=127.5
    ]
 \addplot graphics [includegraphics cmd=\pgfimage,xmin=-0.5, xmax=127.5, ymin=127.5, ymax=-0.5] {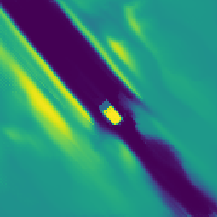};
\end{groupplot}

\begin{groupplot}[
  group style={
      group name=right plots,
      group size=5 by 1,
      horizontal sep=0pt,
      x descriptions at=edge bottom},
  width=1.25cm,
  height=1.25cm,
  scale only axis,
  every axis title/.style={yshift=4pt, xshift=17.7pt}]
   
\nextgroupplot[
anchor=north west, at={($(left plots c5r1.north east) + (0.5cm,0.0cm)$)},
title={$t=-4$},
axis background/.style={fill=white!89.8039215686275!black},
axis line style={black},
xmajorgrids,
xmajorticks=false,
xmin=-0.5, xmax=127.5,
y dir=reverse,
ymajorgrids,
ymajorticks=false,
ymin=-0.5, ymax=127.5
]
\addplot graphics [includegraphics cmd=\pgfimage,xmin=-0.5, xmax=127.5, ymin=127.5, ymax=-0.5] {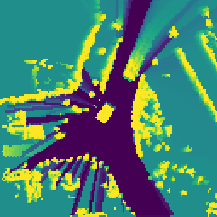};

\nextgroupplot[
title={$t=-3$},
axis background/.style={fill=white!89.8039215686275!black},
axis line style={black},
xmajorgrids,
xmajorticks=false,
xmin=-0.5, xmax=127.5,
y dir=reverse,
ymajorgrids,
ymajorticks=false,
ymin=-0.5, ymax=127.5
]
\addplot graphics [includegraphics cmd=\pgfimage,xmin=-0.5, xmax=127.5, ymin=127.5, ymax=-0.5] {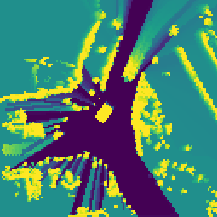};

\nextgroupplot[
title={$t=-2$},
axis background/.style={fill=white!89.8039215686275!black},
axis line style={black},
xmajorgrids,
xmajorticks=false,
xmin=-0.5, xmax=127.5,
y dir=reverse,
ymajorgrids,
ymajorticks=false,
ymin=-0.5, ymax=127.5
]
\addplot graphics [includegraphics cmd=\pgfimage,xmin=-0.5, xmax=127.5, ymin=127.5, ymax=-0.5] {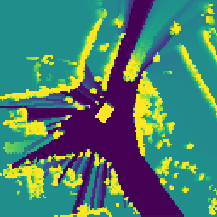};

\nextgroupplot[
title={$t=-1$},
axis background/.style={fill=white!89.8039215686275!black},
axis line style={black},
xmajorgrids,
xmajorticks=false,
xmin=-0.5, xmax=127.5,
y dir=reverse,
ymajorgrids,
ymajorticks=false,
ymin=-0.5, ymax=127.5
]
\addplot graphics [includegraphics cmd=\pgfimage,xmin=-0.5, xmax=127.5, ymin=127.5, ymax=-0.5] {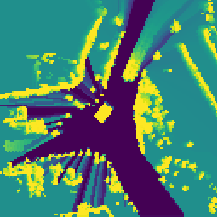};

\nextgroupplot[
title={$t=0$},
axis background/.style={fill=white!89.8039215686275!black},
axis line style={black},
xmajorgrids,
xmajorticks=false,
xmin=-0.5, xmax=127.5,
y dir=reverse,
ymajorgrids,
ymajorticks=false,
ymin=-0.5, ymax=127.5
]
\addplot graphics [includegraphics cmd=\pgfimage,xmin=-0.5, xmax=127.5, ymin=127.5, ymax=-0.5] {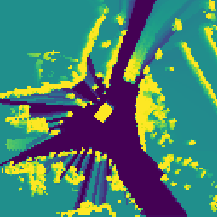};

\end{groupplot}

\begin{groupplot}[
    group style={
      group size=5 by 6,
      horizontal sep=0pt,
      vertical sep = 0pt,
      x descriptions at=edge bottom},
    width=1.25cm,
    height=1.25cm,
    scale only axis,
    ytick pos=left,
    every axis title/.style={yshift=4pt, xshift=17.7pt}]
 \nextgroupplot[
    anchor=north west, at={($(right plots c1r1.south west) - (0.0cm,0.6cm)$)},
    title={$t=5$},
    axis background/.style={fill=white!89.8039215686275!black},
    axis line style={black},
    xmajorgrids,
    xmajorticks=false,
    xmin=-0.5, xmax=127.5,
    y dir=reverse,
    ymajorgrids,
    ymajorticks=false,
    ymin=-0.5, ymax=127.5
    ]
  \addplot graphics [includegraphics cmd=\pgfimage,xmin=-0.5, xmax=127.5, ymin=127.5, ymax=-0.5] {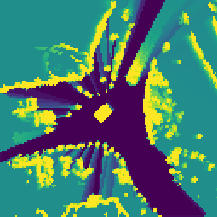};
 
  \nextgroupplot[
    title={$t=10$},
    axis background/.style={fill=white!89.8039215686275!black},
    axis line style={black},
    xmajorgrids,
    xmajorticks=false,
    xmin=-0.5, xmax=127.5,
    y dir=reverse,
    ymajorgrids,
    ymajorticks=false,
    ymin=-0.5, ymax=127.5
    ]
 \addplot graphics [includegraphics cmd=\pgfimage,xmin=-0.5, xmax=127.5, ymin=127.5, ymax=-0.5] {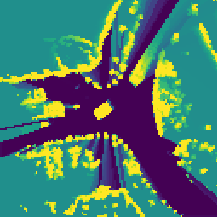};
 
 \nextgroupplot[
    title={$t=15$},
    axis background/.style={fill=white!89.8039215686275!black},
    axis line style={black},
    xmajorgrids,
    xmajorticks=false,
    xmin=-0.5, xmax=127.5,
    y dir=reverse,
    ymajorgrids,
    ymajorticks=false,
    ymin=-0.5, ymax=127.5
    ]
 \addplot graphics [includegraphics cmd=\pgfimage,xmin=-0.5, xmax=127.5, ymin=127.5, ymax=-0.5] {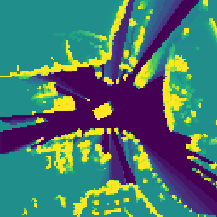};
 \nextgroupplot[
    title={$t=20$},
    axis background/.style={fill=white!89.8039215686275!black},
    axis line style={black},
    xmajorgrids,
    xmajorticks=false,
    xmin=-0.5, xmax=127.5,
    y dir=reverse,
    ymajorgrids,
    ymajorticks=false,
    ymin=-0.5, ymax=127.5
    ]
 \addplot graphics [includegraphics cmd=\pgfimage,xmin=-0.5, xmax=127.5, ymin=127.5, ymax=-0.5] {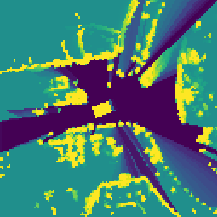};
 \nextgroupplot[
    title={$t=25$},
    axis background/.style={fill=white!89.8039215686275!black},
    axis line style={black},
    xmajorgrids,
    xmajorticks=false,
    xmin=-0.5, xmax=127.5,
    y dir=reverse,
    ymajorgrids,
    ymajorticks=false,
    ymin=-0.5, ymax=127.5
    ]
 \addplot graphics [includegraphics cmd=\pgfimage,xmin=-0.5, xmax=127.5, ymin=127.5, ymax=-0.5] {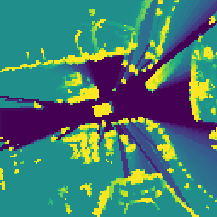};
\nextgroupplot[
    axis background/.style={fill=white!89.8039215686275!black},
    axis line style={black},
    xmajorgrids,
    xmajorticks=false,
    xmin=-0.5, xmax=127.5,
    y dir=reverse,
    ymajorgrids,
    ymajorticks=false,
    ymin=-0.5, ymax=127.5
    ]
 \addplot graphics [includegraphics cmd=\pgfimage,xmin=-0.5, xmax=127.5, ymin=127.5, ymax=-0.5] {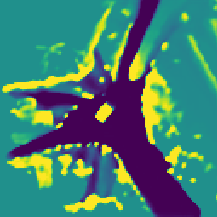};
 \nextgroupplot[
    axis background/.style={fill=white!89.8039215686275!black},
    axis line style={black},
    xmajorgrids,
    xmajorticks=false,
    xmin=-0.5, xmax=127.5,
    y dir=reverse,
    ymajorgrids,
    ymajorticks=false,
    ymin=-0.5, ymax=127.5
    ]
 \addplot graphics [includegraphics cmd=\pgfimage,xmin=-0.5, xmax=127.5, ymin=127.5, ymax=-0.5] {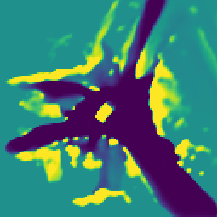};
 \nextgroupplot[
    axis background/.style={fill=white!89.8039215686275!black},
    axis line style={black},
    xmajorgrids,
    xmajorticks=false,
    xmin=-0.5, xmax=127.5,
    y dir=reverse,
    ymajorgrids,
    ymajorticks=false,
    ymin=-0.5, ymax=127.5
    ]
 \addplot graphics [includegraphics cmd=\pgfimage,xmin=-0.5, xmax=127.5, ymin=127.5, ymax=-0.5] {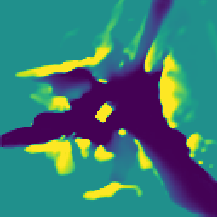};
 \nextgroupplot[
    axis background/.style={fill=white!89.8039215686275!black},
    axis line style={black},
    xmajorgrids,
    xmajorticks=false,
    xmin=-0.5, xmax=127.5,
    y dir=reverse,
    ymajorgrids,
    ymajorticks=false,
    ymin=-0.5, ymax=127.5
    ]
 \addplot graphics [includegraphics cmd=\pgfimage,xmin=-0.5, xmax=127.5, ymin=127.5, ymax=-0.5] {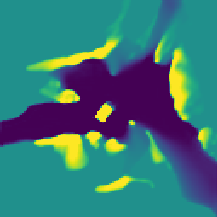};
 \nextgroupplot[
    axis background/.style={fill=white!89.8039215686275!black},
    axis line style={black},
    xmajorgrids,
    xmajorticks=false,
    xmin=-0.5, xmax=127.5,
    y dir=reverse,
    ymajorgrids,
    ymajorticks=false,
    ymin=-0.5, ymax=127.5
    ]
 \addplot graphics [includegraphics cmd=\pgfimage,xmin=-0.5, xmax=127.5, ymin=127.5, ymax=-0.5] {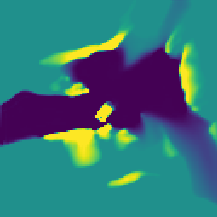};
\nextgroupplot[
    axis background/.style={fill=white!89.8039215686275!black},
    axis line style={black},
    xmajorgrids,
    xmajorticks=false,
    xmin=-0.5, xmax=127.5,
    y dir=reverse,
    ymajorgrids,
    ymajorticks=false,
    ymin=-0.5, ymax=127.5
    ]
 \addplot graphics [includegraphics cmd=\pgfimage,xmin=-0.5, xmax=127.5, ymin=127.5, ymax=-0.5] {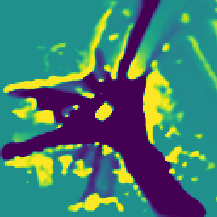};
 \nextgroupplot[
    axis background/.style={fill=white!89.8039215686275!black},
    axis line style={black},
    xmajorgrids,
    xmajorticks=false,
    xmin=-0.5, xmax=127.5,
    y dir=reverse,
    ymajorgrids,
    ymajorticks=false,
    ymin=-0.5, ymax=127.5
    ]
 \addplot graphics [includegraphics cmd=\pgfimage,xmin=-0.5, xmax=127.5, ymin=127.5, ymax=-0.5] {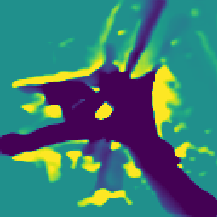};
 \nextgroupplot[
    axis background/.style={fill=white!89.8039215686275!black},
    axis line style={black},
    xmajorgrids,
    xmajorticks=false,
    xmin=-0.5, xmax=127.5,
    y dir=reverse,
    ymajorgrids,
    ymajorticks=false,
    ymin=-0.5, ymax=127.5
    ]
 \addplot graphics [includegraphics cmd=\pgfimage,xmin=-0.5, xmax=127.5, ymin=127.5, ymax=-0.5] {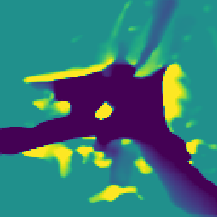};
 \nextgroupplot[
    axis background/.style={fill=white!89.8039215686275!black},
    axis line style={black},
    xmajorgrids,
    xmajorticks=false,
    xmin=-0.5, xmax=127.5,
    y dir=reverse,
    ymajorgrids,
    ymajorticks=false,
    ymin=-0.5, ymax=127.5
    ]
 \addplot graphics [includegraphics cmd=\pgfimage,xmin=-0.5, xmax=127.5, ymin=127.5, ymax=-0.5] {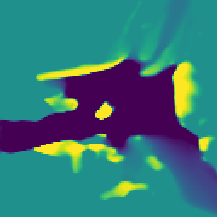};
 \nextgroupplot[
    axis background/.style={fill=white!89.8039215686275!black},
    axis line style={black},
    xmajorgrids,
    xmajorticks=false,
    xmin=-0.5, xmax=127.5,
    y dir=reverse,
    ymajorgrids,
    ymajorticks=false,
    ymin=-0.5, ymax=127.5
    ]
 \addplot graphics [includegraphics cmd=\pgfimage,xmin=-0.5, xmax=127.5, ymin=127.5, ymax=-0.5] {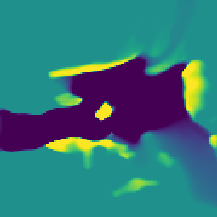};
\nextgroupplot[
    axis background/.style={fill=white!89.8039215686275!black},
    axis line style={black},
    xmajorgrids,
    xmajorticks=false,
    xmin=-0.5, xmax=127.5,
    y dir=reverse,
    ymajorgrids,
    ymajorticks=false,
    ymin=-0.5, ymax=127.5
    ]
 \addplot graphics [includegraphics cmd=\pgfimage,xmin=-0.5, xmax=127.5, ymin=127.5, ymax=-0.5] {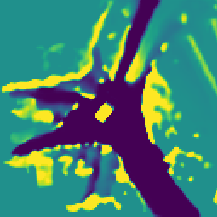};
 \nextgroupplot[
    axis background/.style={fill=white!89.8039215686275!black},
    axis line style={black},
    xmajorgrids,
    xmajorticks=false,
    xmin=-0.5, xmax=127.5,
    y dir=reverse,
    ymajorgrids,
    ymajorticks=false,
    ymin=-0.5, ymax=127.5
    ]
 \addplot graphics [includegraphics cmd=\pgfimage,xmin=-0.5, xmax=127.5, ymin=127.5, ymax=-0.5] {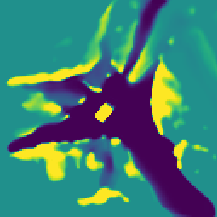};
 \nextgroupplot[
    axis background/.style={fill=white!89.8039215686275!black},
    axis line style={black},
    xmajorgrids,
    xmajorticks=false,
    xmin=-0.5, xmax=127.5,
    y dir=reverse,
    ymajorgrids,
    ymajorticks=false,
    ymin=-0.5, ymax=127.5
    ]
 \addplot graphics [includegraphics cmd=\pgfimage,xmin=-0.5, xmax=127.5, ymin=127.5, ymax=-0.5] {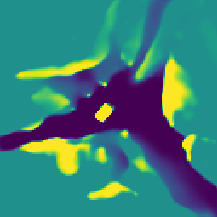};
 \nextgroupplot[
    axis background/.style={fill=white!89.8039215686275!black},
    axis line style={black},
    xmajorgrids,
    xmajorticks=false,
    xmin=-0.5, xmax=127.5,
    y dir=reverse,
    ymajorgrids,
    ymajorticks=false,
    ymin=-0.5, ymax=127.5
    ]
 \addplot graphics [includegraphics cmd=\pgfimage,xmin=-0.5, xmax=127.5, ymin=127.5, ymax=-0.5] {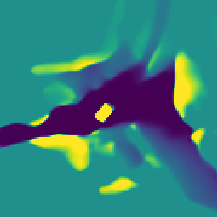};
 \nextgroupplot[
    axis background/.style={fill=white!89.8039215686275!black},
    axis line style={black},
    xmajorgrids,
    xmajorticks=false,
    xmin=-0.5, xmax=127.5,
    y dir=reverse,
    ymajorgrids,
    ymajorticks=false,
    ymin=-0.5, ymax=127.5
    ]
 \addplot graphics [includegraphics cmd=\pgfimage,xmin=-0.5, xmax=127.5, ymin=127.5, ymax=-0.5] {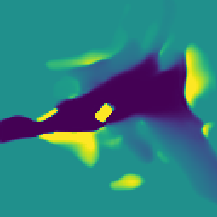};
\nextgroupplot[
    axis background/.style={fill=white!89.8039215686275!black},
    axis line style={black},
    xmajorgrids,
    xmajorticks=false,
    xmin=-0.5, xmax=127.5,
    y dir=reverse,
    ymajorgrids,
    ymajorticks=false,
    ymin=-0.5, ymax=127.5
    ]
 \addplot graphics [includegraphics cmd=\pgfimage,xmin=-0.5, xmax=127.5, ymin=127.5, ymax=-0.5] {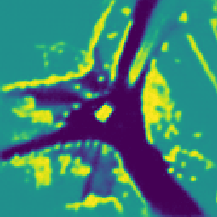};
 
 \nextgroupplot[
    axis background/.style={fill=white!89.8039215686275!black},
    axis line style={black},
    xmajorgrids,
    xmajorticks=false,
    xmin=-0.5, xmax=127.5,
    y dir=reverse,
    ymajorgrids,
    ymajorticks=false,
    ymin=-0.5, ymax=127.5
    ]
 \addplot graphics [includegraphics cmd=\pgfimage,xmin=-0.5, xmax=127.5, ymin=127.5, ymax=-0.5] {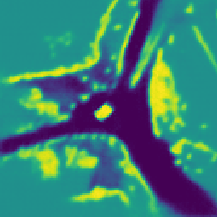};
 
 \nextgroupplot[
    axis background/.style={fill=white!89.8039215686275!black},
    axis line style={black},
    xmajorgrids,
    xmajorticks=false,
    xmin=-0.5, xmax=127.5,
    y dir=reverse,
    ymajorgrids,
    ymajorticks=false,
    ymin=-0.5, ymax=127.5
    ]
 \addplot graphics [includegraphics cmd=\pgfimage,xmin=-0.5, xmax=127.5, ymin=127.5, ymax=-0.5] {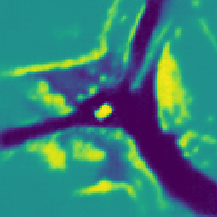};
 
 \nextgroupplot[
    axis background/.style={fill=white!89.8039215686275!black},
    axis line style={black},
    xmajorgrids,
    xmajorticks=false,
    xmin=-0.5, xmax=127.5,
    y dir=reverse,
    ymajorgrids,
    ymajorticks=false,
    ymin=-0.5, ymax=127.5
    ]
 \addplot graphics [includegraphics cmd=\pgfimage,xmin=-0.5, xmax=127.5, ymin=127.5, ymax=-0.5] {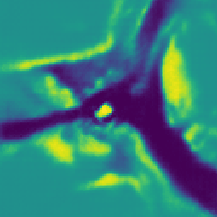};
 
 \nextgroupplot[
    axis background/.style={fill=white!89.8039215686275!black},
    axis line style={black},
    xmajorgrids,
    xmajorticks=false,
    xmin=-0.5, xmax=127.5,
    y dir=reverse,
    ymajorgrids,
    ymajorticks=false,
    ymin=-0.5, ymax=127.5
    ]
 \addplot graphics [includegraphics cmd=\pgfimage,xmin=-0.5, xmax=127.5, ymin=127.5, ymax=-0.5] {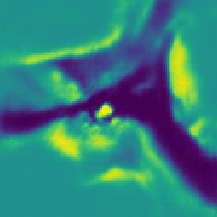};
 
   \nextgroupplot[
    axis background/.style={fill=white!89.8039215686275!black},
    axis line style={black},
    xmajorgrids,
    xmajorticks=false,
    xmin=-0.5, xmax=127.5,
    y dir=reverse,
    ymajorgrids,
    ymajorticks=false,
    ymin=-0.5, ymax=127.5
    ]
 \addplot graphics [includegraphics cmd=\pgfimage,xmin=-0.5, xmax=127.5, ymin=127.5, ymax=-0.5] {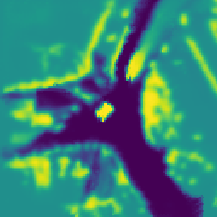};
 
  \nextgroupplot[
    axis background/.style={fill=white!89.8039215686275!black},
    axis line style={black},
    xmajorgrids,
    xmajorticks=false,
    xmin=-0.5, xmax=127.5,
    y dir=reverse,
    ymajorgrids,
    ymajorticks=false,
    ymin=-0.5, ymax=127.5
    ]
 \addplot graphics [includegraphics cmd=\pgfimage,xmin=-0.5, xmax=127.5, ymin=127.5, ymax=-0.5] {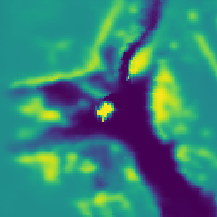};
 
  \nextgroupplot[
    axis background/.style={fill=white!89.8039215686275!black},
    axis line style={black},
    xmajorgrids,
    xmajorticks=false,
    xmin=-0.5, xmax=127.5,
    y dir=reverse,
    ymajorgrids,
    ymajorticks=false,
    ymin=-0.5, ymax=127.5
    ]
 \addplot graphics [includegraphics cmd=\pgfimage,xmin=-0.5, xmax=127.5, ymin=127.5, ymax=-0.5] {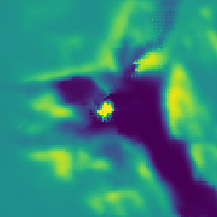};
 
  \nextgroupplot[
    axis background/.style={fill=white!89.8039215686275!black},
    axis line style={black},
    xmajorgrids,
    xmajorticks=false,
    xmin=-0.5, xmax=127.5,
    y dir=reverse,
    ymajorgrids,
    ymajorticks=false,
    ymin=-0.5, ymax=127.5
    ]
 \addplot graphics [includegraphics cmd=\pgfimage,xmin=-0.5, xmax=127.5, ymin=127.5, ymax=-0.5] {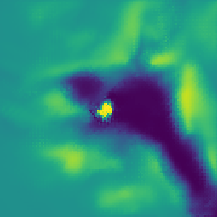};
 
  \nextgroupplot[
    axis background/.style={fill=white!89.8039215686275!black},
    axis line style={black},
    xmajorgrids,
    xmajorticks=false,
    xmin=-0.5, xmax=127.5,
    y dir=reverse,
    ymajorgrids,
    ymajorticks=false,
    ymin=-0.5, ymax=127.5
    ]
 \addplot graphics [includegraphics cmd=\pgfimage,xmin=-0.5, xmax=127.5, ymin=127.5, ymax=-0.5] {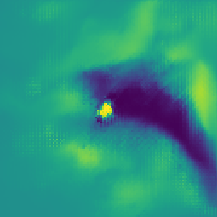};
\end{groupplot}

\begin{groupplot}[
    group style={
      group name=legend,
      group size=1 by 1,
      horizontal sep=0pt,
      vertical sep = 0pt,
      x descriptions at=edge bottom},
    width=5.23cm,
    height=0.5cm,
    scale only axis,
    ytick pos=left,
    every axis title/.style={yshift=4pt, xshift=17.7pt}]
 \nextgroupplot[
    anchor=north west, at={($(left bottom c5r6.south west) - (1cm,0.25cm)$)},
    axis background/.style={fill=white!89.8039215686275!black},
    axis line style={white},
    xmajorgrids,
    xmajorticks=false,
    xmin=-0.5, xmax=127.5,
    y dir=reverse,
    ymajorgrids,
    ymajorticks=false,
    ymin=-0.5, ymax=127.5
    ]
  \addplot graphics [includegraphics cmd=\pgfimage,xmin=-0.5, xmax=127.5, ymin=127.5, ymax=-0.5] {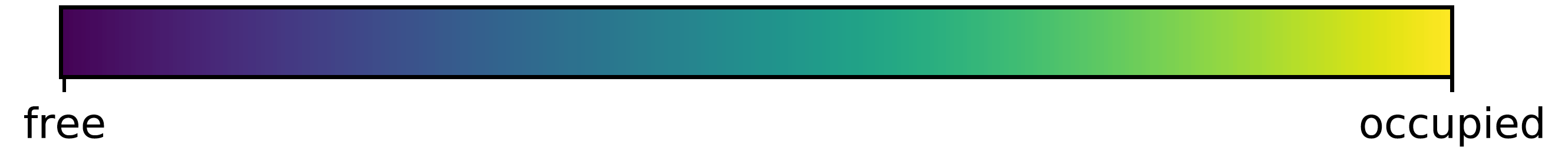};
 \end{groupplot}
\end{tikzpicture}}
\caption{Example \SI{2.5}{\second} predictions on the KITTI dataset and the associated ground truth. Left: The ego-vehicle passes an oncoming vehicle which is preserved in the predictions made by our methods, unlike PredNet, PredRNN++ and CDNA. Right: The ego-vehicle makes a turn at an intersection. Our methods maintain the presence of the vehicle adjacent to the ego-vehicle and better represent the intersection.}
\label{fig:KittiVis}
\end{center}
\end{figure*}

%% file: figures/IS_MOBBM.tex
\begin{tikzpicture}

\definecolor{color0}{rgb}{1,0.647058823529412,0}
\definecolor{color1}{rgb}{0.501960784313725,0,0.501960784313725}

\begin{groupplot}[group style={group size=1 by 2, vertical sep=7pt}, width=1.0\linewidth]

\definecolor{color0}{rgb}{1,0.647058823529412,0}
\definecolor{color1}{rgb}{0.501960784313725,0,0.501960784313725}

\nextgroupplot[
width=\linewidth,
height=4.5cm,
legend cell align={left},
legend style={fill opacity=0.8, draw opacity=1, text opacity=1, at={(0.5,-1.9)}, draw=white!80!black, font={\scriptsize}, legend columns=2, /tikz/every even column/.append style={column sep=0.4cm}, anchor=south},
tick align=outside,
tick pos=left,
x grid style={white!69.0196078431373!black},
xlabel={},
xmin=5, xmax=24,
xtick style={color=black},
ticklabel style = {font=\scriptsize},
y grid style={white!69.0196078431373!black},
ylabel={\small MOBBM},
ymin=0, ymax=0.6,
ytick style={color=black},
ytick={0,0.1,0.2,0.3,0.4,0.5,0.6},
yticklabels={0.0,0.1,0.2,0.3,0.4,0.5,},
xtick={5,10,15,20},
xticklabels={,,,}
]

\path [draw=red, fill=red, opacity=0.2]
(axis cs:1,0.819258101166346)
--(axis cs:1,0.608776695662555)
--(axis cs:2,0.54314693080514)
--(axis cs:3,0.45757535756837)
--(axis cs:4,0.395889495914936)
--(axis cs:5,0.343200052904557)
--(axis cs:6,0.294500792368489)
--(axis cs:7,0.269199614013653)
--(axis cs:8,0.243152020961165)
--(axis cs:9,0.219153939000259)
--(axis cs:10,0.203978564525392)
--(axis cs:11,0.1931580249639)
--(axis cs:12,0.207223373725862)
--(axis cs:13,0.233334356246771)
--(axis cs:14,0.264257167675301)
--(axis cs:15,0.28735552926918)
--(axis cs:16,0.313128554015402)
--(axis cs:17,0.342459441494281)
--(axis cs:18,0.350142650176131)
--(axis cs:19,0.329794706326916)
--(axis cs:20,0.282436293500382)
--(axis cs:21,0.217006873061181)
--(axis cs:22,0.146196643163332)
--(axis cs:23,0.0969705698196203)
--(axis cs:24,0.061985535461908)
--(axis cs:24,0.101426629415697)
--(axis cs:24,0.101426629415697)
--(axis cs:23,0.155005500502413)
--(axis cs:22,0.218618329777958)
--(axis cs:21,0.314992162251965)
--(axis cs:20,0.406900787555999)
--(axis cs:19,0.469219238941083)
--(axis cs:18,0.504442648991069)
--(axis cs:17,0.499493744463985)
--(axis cs:16,0.467693611270265)
--(axis cs:15,0.431526543631037)
--(axis cs:14,0.397654767641438)
--(axis cs:13,0.356247633785472)
--(axis cs:12,0.317878535091395)
--(axis cs:11,0.296240437762829)
--(axis cs:10,0.308195054147153)
--(axis cs:9,0.327103183553)
--(axis cs:8,0.357698955487522)
--(axis cs:7,0.387839421689117)
--(axis cs:6,0.418602626520792)
--(axis cs:5,0.476592374323333)
--(axis cs:4,0.544321083907537)
--(axis cs:3,0.620823928997145)
--(axis cs:2,0.732457846255252)
--(axis cs:1,0.819258101166346)
--cycle;

\addplot [semithick, red, mark=triangle*, mark size=3, mark options={solid}]
table {%
1 0.714017398414451
2 0.637802388530196
3 0.539199643282758
4 0.470105289911237
5 0.409896213613945
6 0.35655170944464
7 0.328519517851385
8 0.300425488224344
9 0.27312856127663
10 0.256086809336272
11 0.244699231363365
12 0.262550954408628
13 0.294790995016122
14 0.330955967658369
15 0.359441036450108
16 0.390411082642834
17 0.420976592979133
18 0.4272926495836
19 0.399506972634
20 0.344668540528191
21 0.265999517656573
22 0.182407486470645
23 0.125988035161017
24 0.0817060824388027
};
\addlegendentry{TAAConvLSTM (ours)} 

\path [draw=green!66!black, fill=green!66!black, opacity=0.2]
(axis cs:1,0.788732423204761)
--(axis cs:1,0.588966656431889)
--(axis cs:2,0.541238371340658)
--(axis cs:3,0.473799106547029)
--(axis cs:4,0.420857911377351)
--(axis cs:5,0.374675250152433)
--(axis cs:6,0.333299564039507)
--(axis cs:7,0.299361670321247)
--(axis cs:8,0.264742495461903)
--(axis cs:9,0.242237632856707)
--(axis cs:10,0.207277255338445)
--(axis cs:11,0.179169718905461)
--(axis cs:12,0.159288031999253)
--(axis cs:13,0.145332099411563)
--(axis cs:14,0.143152500221426)
--(axis cs:15,0.154212037249715)
--(axis cs:16,0.172870502932865)
--(axis cs:17,0.202878974439426)
--(axis cs:18,0.220510463690884)
--(axis cs:19,0.230682649952491)
--(axis cs:20,0.236984424803643)
--(axis cs:21,0.218348186828905)
--(axis cs:22,0.192107840523777)
--(axis cs:23,0.159681432041515)
--(axis cs:24,0.128255480828416)
--(axis cs:24,0.202410773631278)
--(axis cs:24,0.202410773631278)
--(axis cs:23,0.249831673482009)
--(axis cs:22,0.295471035023483)
--(axis cs:21,0.333525093217639)
--(axis cs:20,0.358272191408339)
--(axis cs:19,0.350084943845395)
--(axis cs:18,0.338819043503074)
--(axis cs:17,0.311401929258751)
--(axis cs:16,0.267056998736343)
--(axis cs:15,0.240507799058087)
--(axis cs:14,0.222831105408645)
--(axis cs:13,0.227544979874562)
--(axis cs:12,0.24623025692147)
--(axis cs:11,0.269117001552711)
--(axis cs:10,0.304098632067389)
--(axis cs:9,0.347892664301088)
--(axis cs:8,0.376199278787562)
--(axis cs:7,0.417921975165482)
--(axis cs:6,0.461641454464405)
--(axis cs:5,0.513051322881244)
--(axis cs:4,0.571648870614194)
--(axis cs:3,0.637382477153944)
--(axis cs:2,0.724434874523447)
--(axis cs:1,0.788732423204761)
--cycle;

\addplot [semithick, green!66!black, mark=x, mark size=3, mark options={solid}]
table {%
1 0.688849539818325
2 0.632836622932052
3 0.555590791850486
4 0.496253390995772
5 0.443863286516838
6 0.397470509251956
7 0.358641822743365
8 0.320470887124732
9 0.295065148578898
10 0.255687943702917
11 0.224143360229086
12 0.202759144460361
13 0.186438539643062
14 0.182991802815036
15 0.197359918153901
16 0.219963750834604
17 0.257140451849088
18 0.279664753596979
19 0.290383796898943
20 0.297628308105991
21 0.275936640023272
22 0.24378943777363
23 0.204756552761762
24 0.165333127229847
};
\addlegendentry{SAAConvLSTM (ours)} 

\path [draw=color0, fill=color0, opacity=0.2]
(axis cs:1,0.858769073345935)
--(axis cs:1,0.638461490398501)
--(axis cs:2,0.612054830578217)
--(axis cs:3,0.52848907852192)
--(axis cs:4,0.472731401693919)
--(axis cs:5,0.405260712818479)
--(axis cs:6,0.367762722989414)
--(axis cs:7,0.324254259124048)
--(axis cs:8,0.304063381338259)
--(axis cs:9,0.266776524375119)
--(axis cs:10,0.247509567700861)
--(axis cs:11,0.226146391452913)
--(axis cs:12,0.200628912945636)
--(axis cs:13,0.181863988251508)
--(axis cs:14,0.16096553920747)
--(axis cs:15,0.150631615247022)
--(axis cs:16,0.142191885582785)
--(axis cs:17,0.137111640483119)
--(axis cs:18,0.138014539513514)
--(axis cs:19,0.139887349402495)
--(axis cs:20,0.141567232832173)
--(axis cs:21,0.137768120347018)
--(axis cs:22,0.130781781707554)
--(axis cs:23,0.118503279180023)
--(axis cs:24,0.103627726852656)
--(axis cs:24,0.16050173171359)
--(axis cs:24,0.16050173171359)
--(axis cs:23,0.182234327047414)
--(axis cs:22,0.201181409871609)
--(axis cs:21,0.212320123762264)
--(axis cs:20,0.218250809399035)
--(axis cs:19,0.214366750344734)
--(axis cs:18,0.21085713711629)
--(axis cs:17,0.207419257143503)
--(axis cs:16,0.211375189043683)
--(axis cs:15,0.221479218086189)
--(axis cs:14,0.237293571457361)
--(axis cs:13,0.261324723421723)
--(axis cs:12,0.28720982783938)
--(axis cs:11,0.320785565036046)
--(axis cs:10,0.349230237421522)
--(axis cs:9,0.377265838808625)
--(axis cs:8,0.426373564112059)
--(axis cs:7,0.451606368386789)
--(axis cs:6,0.504955376872806)
--(axis cs:5,0.5551379814856)
--(axis cs:4,0.641062253816134)
--(axis cs:3,0.712693984662901)
--(axis cs:2,0.820650627683313)
--(axis cs:1,0.858769073345935)
--cycle;

\addplot [semithick, color0, mark=asterisk, mark size=3, mark options={solid}]
table {%
1 0.748615281872218
2 0.716352729130765
3 0.62059153159241
4 0.556896827755026
5 0.48019934715204
6 0.43635904993111
7 0.387930313755418
8 0.365218472725159
9 0.322021181591872
10 0.298369902561191
11 0.27346597824448
12 0.243919370392508
13 0.221594355836616
14 0.199129555332416
15 0.186055416666606
16 0.176783537313234
17 0.172265448813311
18 0.174435838314902
19 0.177127049873614
20 0.179909021115604
21 0.175044122054641
22 0.165981595789581
23 0.150368803113719
24 0.132064729283123
};
\addlegendentry{PredNet}

\path [draw=white!50.1960784313725!black, fill=white!50.1960784313725!black, opacity=0.2]
(axis cs:1,0.257095466587085)
--(axis cs:1,0.155463729928404)
--(axis cs:2,0.167470526195099)
--(axis cs:3,0.169861026157038)
--(axis cs:4,0.140957770747823)
--(axis cs:5,0.135754273411909)
--(axis cs:6,0.1290434991064)
--(axis cs:7,0.119872663602334)
--(axis cs:8,0.131001968834822)
--(axis cs:9,0.142363771340191)
--(axis cs:10,0.117050526699704)
--(axis cs:11,0.0895783052326148)
--(axis cs:12,0.0775708198674218)
--(axis cs:13,0.079220077552441)
--(axis cs:14,0.053435927626876)
--(axis cs:15,0.0391154857241227)
--(axis cs:16,0.0350172302079972)
--(axis cs:17,0.0376976101251927)
--(axis cs:18,0.0446681359530752)
--(axis cs:19,0.0469608873943871)
--(axis cs:20,0.0529163106289917)
--(axis cs:21,0.0653300942105099)
--(axis cs:22,0.0771834566640365)
--(axis cs:23,0.100808695380349)
--(axis cs:24,0.0871275349356645)
--(axis cs:24,0.151224111671377)
--(axis cs:24,0.151224111671377)
--(axis cs:23,0.172910146209455)
--(axis cs:22,0.138156710786226)
--(axis cs:21,0.1256785901478)
--(axis cs:20,0.106272231273716)
--(axis cs:19,0.0998908860166367)
--(axis cs:18,0.0878649470836139)
--(axis cs:17,0.0744771771765994)
--(axis cs:16,0.0673275720566048)
--(axis cs:15,0.0745777415664798)
--(axis cs:14,0.106298786476197)
--(axis cs:13,0.157173873217662)
--(axis cs:12,0.156655958061014)
--(axis cs:11,0.159802640437076)
--(axis cs:10,0.207656978690675)
--(axis cs:9,0.241810567266328)
--(axis cs:8,0.231370050874845)
--(axis cs:7,0.214586676007134)
--(axis cs:6,0.226988492406951)
--(axis cs:5,0.239557609283713)
--(axis cs:4,0.242840666817131)
--(axis cs:3,0.282246574201643)
--(axis cs:2,0.271831959040043)
--(axis cs:1,0.257095466587085)
--cycle;

\addplot [semithick, white!50.1960784313725!black]
table {%
1 0.206279598257745
2 0.219651242617571
3 0.22605380017934
4 0.191899218782477
5 0.187655941347811
6 0.178015995756675
7 0.167229669804734
8 0.181186009854833
9 0.192087169303259
10 0.162353752695189
11 0.124690472834845
12 0.117113388964218
13 0.118196975385052
14 0.0798673570515367
15 0.0568466136453013
16 0.051172401132301
17 0.056087393650896
18 0.0662665415183445
19 0.0734258867055119
20 0.0795942709513537
21 0.0955043421791551
22 0.107670083725131
23 0.136859420794902
24 0.119175823303521
};
\addlegendentry{CDNA}

\path [draw=color1, fill=color1, opacity=0.2]
(axis cs:1,0.697648260663167)
--(axis cs:1,0.510496516289596)
--(axis cs:2,0.410530908658752)
--(axis cs:3,0.35346120358995)
--(axis cs:4,0.319907467147443)
--(axis cs:5,0.303296711598186)
--(axis cs:6,0.280306390610482)
--(axis cs:7,0.262541720357949)
--(axis cs:8,0.241529863112223)
--(axis cs:9,0.214736216049073)
--(axis cs:10,0.188108385074162)
--(axis cs:11,0.169007318279944)
--(axis cs:12,0.159020460301638)
--(axis cs:13,0.151699774851384)
--(axis cs:14,0.150321223069395)
--(axis cs:15,0.15815728003778)
--(axis cs:16,0.167827614960151)
--(axis cs:17,0.186616105579907)
--(axis cs:18,0.198180862406473)
--(axis cs:19,0.210205810719006)
--(axis cs:20,0.218375319747329)
--(axis cs:21,0.196011468216758)
--(axis cs:22,0.151076148675342)
--(axis cs:23,0.102820979057128)
--(axis cs:24,0.0573126452961774)
--(axis cs:24,0.1049038110199)
--(axis cs:24,0.1049038110199)
--(axis cs:23,0.187995827919889)
--(axis cs:22,0.252270306271703)
--(axis cs:21,0.32137423277068)
--(axis cs:20,0.352576897931977)
--(axis cs:19,0.339923125257302)
--(axis cs:18,0.322629518710289)
--(axis cs:17,0.30899183353966)
--(axis cs:16,0.281493128549103)
--(axis cs:15,0.268578619022691)
--(axis cs:14,0.257800009887635)
--(axis cs:13,0.255556980290918)
--(axis cs:12,0.264392183537159)
--(axis cs:11,0.276178613438219)
--(axis cs:10,0.298294173249278)
--(axis cs:9,0.332686432286491)
--(axis cs:8,0.366213928053688)
--(axis cs:7,0.391005200682046)
--(axis cs:6,0.411141674451817)
--(axis cs:5,0.437267681751409)
--(axis cs:4,0.456622379341687)
--(axis cs:3,0.499110655914679)
--(axis cs:2,0.572980018366351)
--(axis cs:1,0.697648260663167)
--cycle;

\addplot [semithick, color1, mark=+, mark size=3, mark options={solid}]
table {%
1 0.604072388476381
2 0.491755463512552
3 0.426285929752315
4 0.388264923244565
5 0.370282196674798
6 0.345724032531149
7 0.326773460519997
8 0.303871895582956
9 0.273711324167782
10 0.24320127916172
11 0.222592965859081
12 0.211706321919398
13 0.203628377571151
14 0.204060616478515
15 0.213367949530235
16 0.224660371754627
17 0.247803969559784
18 0.260405190558381
19 0.275064467988154
20 0.285476108839653
21 0.258692850493719
22 0.201673227473522
23 0.145408403488509
24 0.0811082281580388
};
\addlegendentry{PredRNN++}

\nextgroupplot[
width=\linewidth,
height=4.5cm,
tick align=outside,
tick pos=left,
x grid style={white!69.0196078431373!black},
xlabel={\small Time Steps},
xmin=5, xmax=24,
ticklabel style = {font=\scriptsize},
xtick style={color=black},
y grid style={white!69.0196078431373!black},
ylabel={\small IS},
ymin=5, ymax=32,
ytick style={color=black}
]

\path [draw=red, fill=red, opacity=0.2]
(axis cs:1,1.89447783623823)
--(axis cs:1,1.81907007109524)
--(axis cs:2,2.44223103199743)
--(axis cs:3,3.23257857230245)
--(axis cs:4,4.07595548330975)
--(axis cs:5,4.95452502775932)
--(axis cs:6,5.82111752001513)
--(axis cs:7,6.75541723325966)
--(axis cs:8,7.65496130062388)
--(axis cs:9,8.68121485674306)
--(axis cs:10,9.66135332684356)
--(axis cs:11,10.6187004144163)
--(axis cs:12,11.6767698910154)
--(axis cs:13,12.7807568888274)
--(axis cs:14,13.8193677478788)
--(axis cs:15,14.8306684549727)
--(axis cs:16,15.8146428288978)
--(axis cs:17,16.7942819997788)
--(axis cs:18,17.8021418633307)
--(axis cs:19,18.7567665663462)
--(axis cs:20,19.7679475296905)
--(axis cs:21,20.7107173891591)
--(axis cs:22,21.6747120909269)
--(axis cs:23,22.5786454448045)
--(axis cs:24,23.5410000297154)
--(axis cs:24,24.5878526028272)
--(axis cs:24,24.5878526028272)
--(axis cs:23,23.5725603628887)
--(axis cs:22,22.6247525915709)
--(axis cs:21,21.6118294496224)
--(axis cs:20,20.6229233074828)
--(axis cs:19,19.5570028732649)
--(axis cs:18,18.5514610823796)
--(axis cs:17,17.4898594270081)
--(axis cs:16,16.4583143905128)
--(axis cs:15,15.4233918851401)
--(axis cs:14,14.3618069206958)
--(axis cs:13,13.2730926983299)
--(axis cs:12,12.1317663335963)
--(axis cs:11,11.0343747131544)
--(axis cs:10,10.0422326111675)
--(axis cs:9,9.03034427849724)
--(axis cs:8,7.96729092113)
--(axis cs:7,7.04315209321994)
--(axis cs:6,6.07808450640413)
--(axis cs:5,5.18523848386589)
--(axis cs:4,4.27111439788467)
--(axis cs:3,3.38697255516829)
--(axis cs:2,2.56011624370809)
--(axis cs:1,1.89447783623823)
--cycle;

\addplot [semithick, red, mark=triangle*, mark size=3, mark options={solid}]
table {%
1 1.85677395366674
2 2.50117363785276
3 3.30977556373537
4 4.17353494059721
5 5.0698817558126
6 5.94960101320963
7 6.8992846632398
8 7.81112611087694
9 8.85577956762015
10 9.85179296900552
11 10.8265375637854
12 11.9042681123058
13 13.0269247935786
14 14.0905873342873
15 15.1270301700564
16 16.1364786097053
17 17.1420707133935
18 18.1768014728552
19 19.1568847198055
20 20.1954354185867
21 21.1612734193907
22 22.1497323412489
23 23.0756029038466
24 24.0644263162713
};

\path [draw=green!66!black, fill=green!66!black, opacity=0.2]
(axis cs:1,2.02494383282813)
--(axis cs:1,1.94445570816003)
--(axis cs:2,2.48807968555631)
--(axis cs:3,3.4091706438358)
--(axis cs:4,4.27191972920592)
--(axis cs:5,5.22247475310661)
--(axis cs:6,6.21866042651373)
--(axis cs:7,7.23530223768028)
--(axis cs:8,8.39217468633374)
--(axis cs:9,9.43050521217952)
--(axis cs:10,10.5046767252457)
--(axis cs:11,11.5354965232677)
--(axis cs:12,12.5854767480475)
--(axis cs:13,13.5753437756619)
--(axis cs:14,14.6076063791627)
--(axis cs:15,15.6212742057229)
--(axis cs:16,16.6523705356825)
--(axis cs:17,17.7146662785806)
--(axis cs:18,18.7815333430729)
--(axis cs:19,19.882190210626)
--(axis cs:20,20.9870270836208)
--(axis cs:21,22.1048990203854)
--(axis cs:22,23.1856422936672)
--(axis cs:23,24.4148693897835)
--(axis cs:24,25.5810824347039)
--(axis cs:24,26.6635147831054)
--(axis cs:24,26.6635147831054)
--(axis cs:23,25.4259310527659)
--(axis cs:22,24.1134380390616)
--(axis cs:21,22.969439258752)
--(axis cs:20,21.7970208017166)
--(axis cs:19,20.6405365141878)
--(axis cs:18,19.4844668251292)
--(axis cs:17,18.3717548346847)
--(axis cs:16,17.263595843771)
--(axis cs:15,16.1860336768238)
--(axis cs:14,15.1284176383491)
--(axis cs:13,14.0535669555011)
--(axis cs:12,13.0337879721652)
--(axis cs:11,11.9482765371241)
--(axis cs:10,10.8900078800591)
--(axis cs:9,9.7859149777036)
--(axis cs:8,8.72042195328108)
--(axis cs:7,7.53030031891038)
--(axis cs:6,6.48587651677797)
--(axis cs:5,5.46108669462788)
--(axis cs:4,4.48071124734284)
--(axis cs:3,3.57445301112914)
--(axis cs:2,2.6098925460026)
--(axis cs:1,2.02494383282813)
--cycle;

\addplot [semithick, green!66!black, mark=x, mark size=3, mark options={solid}]
table {%
1 1.98469977049408
2 2.54898611577945
3 3.49181182748247
4 4.37631548827438
5 5.34178072386725
6 6.35226847164585
7 7.38280127829533
8 8.55629831980741
9 9.60821009494156
10 10.6973423026524
11 11.7418865301959
12 12.8096323601063
13 13.8144553655815
14 14.8680120087559
15 15.9036539412733
16 16.9579831897268
17 18.0432105566327
18 19.133000084101
19 20.2613633624069
20 21.3920239426687
21 22.5371691395687
22 23.6495401663644
23 24.9204002212747
24 26.1222986089047
};

\path [draw=color0, fill=color0, opacity=0.2]
(axis cs:1,1.94041924210029)
--(axis cs:1,1.87647787689305)
--(axis cs:2,2.52248413366072)
--(axis cs:3,3.42226113099423)
--(axis cs:4,4.34039372217821)
--(axis cs:5,5.40155497632473)
--(axis cs:6,6.44616670819585)
--(axis cs:7,7.49990823173597)
--(axis cs:8,8.71095517346108)
--(axis cs:9,9.91926385313576)
--(axis cs:10,11.0936466981903)
--(axis cs:11,12.1690613745641)
--(axis cs:12,13.2172397819317)
--(axis cs:13,14.1756507501829)
--(axis cs:14,15.150903832376)
--(axis cs:15,16.1359103650044)
--(axis cs:16,17.0828804882494)
--(axis cs:17,18.0970012940923)
--(axis cs:18,19.1197794796164)
--(axis cs:19,20.1581869835928)
--(axis cs:20,21.2196825521986)
--(axis cs:21,22.327685460491)
--(axis cs:22,23.463814723392)
--(axis cs:23,24.5557836517475)
--(axis cs:24,25.6624779635857)
--(axis cs:24,26.7510623746069)
--(axis cs:24,26.7510623746069)
--(axis cs:23,25.5916758978281)
--(axis cs:22,24.4461113658989)
--(axis cs:21,23.2493647353283)
--(axis cs:20,22.0728538794603)
--(axis cs:19,20.9484885078503)
--(axis cs:18,19.8526645963374)
--(axis cs:17,18.7671052448736)
--(axis cs:16,17.7018951218399)
--(axis cs:15,16.7040891910307)
--(axis cs:14,15.6664394683274)
--(axis cs:13,14.6463137712637)
--(axis cs:12,13.6399873563855)
--(axis cs:11,12.5550138742351)
--(axis cs:10,11.4443542923672)
--(axis cs:9,10.2481185235191)
--(axis cs:8,9.01039032301842)
--(axis cs:7,7.7685624748999)
--(axis cs:6,6.69468716553824)
--(axis cs:5,5.627398280694)
--(axis cs:4,4.53269697779236)
--(axis cs:3,3.57586076586289)
--(axis cs:2,2.6314615931723)
--(axis cs:1,1.94041924210029)
--cycle;

\addplot [semithick, color0, mark=asterisk, mark size=3, mark options={solid}]
table {%
1 1.90844855949667
2 2.57697286341651
3 3.49906094842856
4 4.43654534998529
5 5.51447662850936
6 6.57042693686704
7 7.63423535331794
8 8.86067274823975
9 10.0836911883274
10 11.2690004952787
11 12.3620376243996
12 13.4286135691586
13 14.4109822607233
14 15.4086716503517
15 16.4199997780175
16 17.3923878050447
17 18.432053269483
18 19.4862220379769
19 20.5533377457216
20 21.6462682158295
21 22.7885250979096
22 23.9549630446455
23 25.0737297747878
24 26.2067701690963
};

\path [draw=color1, fill=color1, opacity=0.2]
(axis cs:1,2.13868308446778)
--(axis cs:1,2.06776741731049)
--(axis cs:2,2.93202581321827)
--(axis cs:3,3.9144869143886)
--(axis cs:4,5.02083311245296)
--(axis cs:5,6.19692468150889)
--(axis cs:6,7.44638987773716)
--(axis cs:7,8.77325699681181)
--(axis cs:8,10.1050908049179)
--(axis cs:9,11.3153152252681)
--(axis cs:10,12.545828063438)
--(axis cs:11,13.6426700444683)
--(axis cs:12,14.7529890958211)
--(axis cs:13,15.9967253817331)
--(axis cs:14,17.2855103271019)
--(axis cs:15,18.5096926041228)
--(axis cs:16,19.8126623679501)
--(axis cs:17,21.0941527983108)
--(axis cs:18,22.3333408123168)
--(axis cs:19,23.6525792949375)
--(axis cs:20,25.1031661642872)
--(axis cs:21,26.6376928462383)
--(axis cs:22,28.1800943510102)
--(axis cs:23,29.6543811953517)
--(axis cs:24,31.1577659537419)
--(axis cs:24,31.9727811616788)
--(axis cs:24,31.9727811616788)
--(axis cs:23,30.4298137107214)
--(axis cs:22,28.9304530982747)
--(axis cs:21,27.3773514763643)
--(axis cs:20,25.8208202513614)
--(axis cs:19,24.3482275977678)
--(axis cs:18,23.0102357074191)
--(axis cs:17,21.7461760991792)
--(axis cs:16,20.4554594426512)
--(axis cs:15,19.1181715986389)
--(axis cs:14,17.8815864300892)
--(axis cs:13,16.552373684553)
--(axis cs:12,15.2864169302321)
--(axis cs:11,14.1307671490181)
--(axis cs:10,13.0086103767334)
--(axis cs:9,11.7216596813901)
--(axis cs:8,10.484461834363)
--(axis cs:7,9.11296802634424)
--(axis cs:6,7.74404494425942)
--(axis cs:5,6.44604319486789)
--(axis cs:4,5.22988355859373)
--(axis cs:3,4.07233607515445)
--(axis cs:2,3.04522699587166)
--(axis cs:1,2.13868308446778)
--cycle;

\addplot [semithick, color1, mark=+, mark size=3, mark options={solid}]
table {%
1 2.10322525088914
2 2.98862640454496
3 3.99341149477152
4 5.12535833552334
5 6.32148393818839
6 7.59521741099829
7 8.94311251157803
8 10.2947763196405
9 11.5184874533291
10 12.7772192200857
11 13.8867185967432
12 15.0197030130266
13 16.2745495331431
14 17.5835483785956
15 18.8139321013808
16 20.1340609053007
17 21.420164448745
18 22.6717882598679
19 24.0004034463527
20 25.4619932078243
21 27.0075221613013
22 28.5552737246424
23 30.0420974530366
24 31.5652735577104
};

\end{groupplot}

\end{tikzpicture}

%% file: figures/plot_Waymo.tex
\begin{figure*}[h!]
\begin{center}
\scalebox{0.85}{
\begin{tikzpicture} [font=\tiny]
\begin{groupplot}[
   group style={
      group name=left plots,
      group size=5 by 1,
      horizontal sep=0pt,
      x descriptions at=edge bottom},
   width=1.25cm,
   height=1.25cm,
   scale only axis,
   every axis title/.style={yshift=4pt, xshift=17.7pt}]

\nextgroupplot[
title={$t=-4$},
ylabel={\scalebox{.8}{Input}},
axis background/.style={fill=white!89.8039215686275!black},
axis line style={black},
xmajorgrids,
xmajorticks=false,
xmin=-0.5, xmax=127.5,
y dir=reverse,
ymajorgrids,
ymajorticks=false,
ymin=-0.5, ymax=127.5
]
\addplot graphics [includegraphics cmd=\pgfimage,xmin=-0.5, xmax=127.5, ymin=127.5, ymax=-0.5] {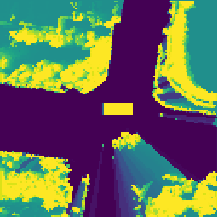};   
   
\nextgroupplot[
title={$t=-3$},
axis background/.style={fill=white!89.8039215686275!black},
axis line style={black},
xmajorgrids,
xmajorticks=false,
xmin=-0.5, xmax=127.5,
y dir=reverse,
ymajorgrids,
ymajorticks=false,
ymin=-0.5, ymax=127.5
]
\addplot graphics [includegraphics cmd=\pgfimage,xmin=-0.5, xmax=127.5, ymin=127.5, ymax=-0.5] {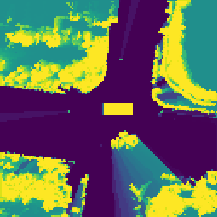};

\nextgroupplot[
title={$t=-2$},
axis background/.style={fill=white!89.8039215686275!black},
axis line style={black},
xmajorgrids,
xmajorticks=false,
xmin=-0.5, xmax=127.5,
y dir=reverse,
ymajorgrids,
ymajorticks=false,
ymin=-0.5, ymax=127.5
]
\addplot graphics [includegraphics cmd=\pgfimage,xmin=-0.5, xmax=127.5, ymin=127.5, ymax=-0.5] {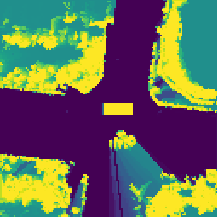};

\nextgroupplot[
title={$t=-1$},
axis background/.style={fill=white!89.8039215686275!black},
axis line style={black},
xmajorgrids,
xmajorticks=false,
xmin=-0.5, xmax=127.5,
y dir=reverse,
ymajorgrids,
ymajorticks=false,
ymin=-0.5, ymax=127.5
]
\addplot graphics [includegraphics cmd=\pgfimage,xmin=-0.5, xmax=127.5, ymin=127.5, ymax=-0.5] {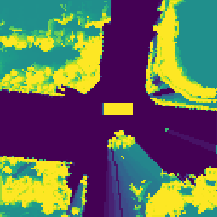};

\nextgroupplot[
title={$t=0$},
axis background/.style={fill=white!89.8039215686275!black},
axis line style={black},
xmajorgrids,
xmajorticks=false,
xmin=-0.5, xmax=127.5,
y dir=reverse,
ymajorgrids,
ymajorticks=false,
ymin=-0.5, ymax=127.5
]
\addplot graphics [includegraphics cmd=\pgfimage,xmin=-0.5, xmax=127.5, ymin=127.5, ymax=-0.5] {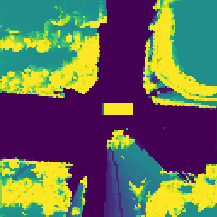};
\end{groupplot}

\begin{groupplot}[
    group style={
        group name=left bottom,
       group size= 5 by 6,
       horizontal sep=0pt,
       vertical sep = 0pt,
       x descriptions at=edge bottom},
    width=1.25cm,
    height=1.25cm,
    scale only axis,
    ytick pos=left,
    every axis title/.style={yshift=4pt, xshift=17.7pt}]
 \nextgroupplot[
    anchor=north west, at={($(left plots c1r1.south west) - (0.0cm,0.6cm)$)},
    title={$t=5$},
    ylabel={\scalebox{.8}{Ground truth}},
    axis background/.style={fill=white!89.8039215686275!black},
    axis line style={black},
    xmajorgrids,
    xmajorticks=false,
    xmin=-0.5, xmax=127.5,
    y dir=reverse,
    ymajorgrids,
    ymajorticks=false,
    ymin=-0.5, ymax=127.5
    ]
 \addplot graphics [includegraphics cmd=\pgfimage,xmin=-0.5, xmax=127.5, ymin=127.5, ymax=-0.5] {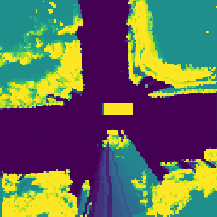};
 
  \nextgroupplot[
    title={$t=10$},
    axis background/.style={fill=white!89.8039215686275!black},
    axis line style={black},
    xmajorgrids,
    xmajorticks=false,
    xmin=-0.5, xmax=127.5,
    y dir=reverse,
    ymajorgrids,
    ymajorticks=false,
    ymin=-0.5, ymax=127.5
    ]
 \addplot graphics [includegraphics cmd=\pgfimage,xmin=-0.5, xmax=127.5, ymin=127.5, ymax=-0.5] {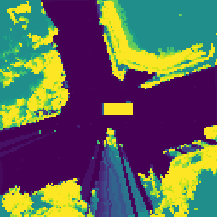};
 
 \nextgroupplot[
    title={$t=15$},
    axis background/.style={fill=white!89.8039215686275!black},
    axis line style={black},
    xmajorgrids,
    xmajorticks=false,
    xmin=-0.5, xmax=127.5,
    y dir=reverse,
    ymajorgrids,
    ymajorticks=false,
    ymin=-0.5, ymax=127.5
    ]
 \addplot graphics [includegraphics cmd=\pgfimage,xmin=-0.5, xmax=127.5, ymin=127.5, ymax=-0.5] {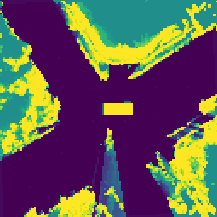};
 \nextgroupplot[
    title={$t=20$},
    axis background/.style={fill=white!89.8039215686275!black},
    axis line style={black},
    xmajorgrids,
    xmajorticks=false,
    xmin=-0.5, xmax=127.5,
    y dir=reverse,
    ymajorgrids,
    ymajorticks=false,
    ymin=-0.5, ymax=127.5
    ]
 \addplot graphics [includegraphics cmd=\pgfimage,xmin=-0.5, xmax=127.5, ymin=127.5, ymax=-0.5] {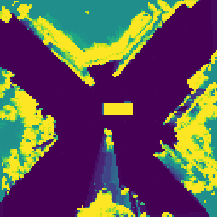};
 \nextgroupplot[
    title={$t=25$},
    axis background/.style={fill=white!89.8039215686275!black},
    axis line style={black},
    xmajorgrids,
    xmajorticks=false,
    xmin=-0.5, xmax=127.5,
    y dir=reverse,
    ymajorgrids,
    ymajorticks=false,
    ymin=-0.5, ymax=127.5
    ]
 \addplot graphics [includegraphics cmd=\pgfimage,xmin=-0.5, xmax=127.5, ymin=127.5, ymax=-0.5] {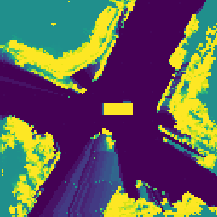};
\nextgroupplot[
    ylabel={\scalebox{.8}{TAAConvLSTM}},
    axis background/.style={fill=white!89.8039215686275!black},
    axis line style={black},
    xmajorgrids,
    xmajorticks=false,
    xmin=-0.5, xmax=127.5,
    y dir=reverse,
    ymajorgrids,
    ymajorticks=false,
    ymin=-0.5, ymax=127.5
    ]
 \addplot graphics [includegraphics cmd=\pgfimage,xmin=-0.5, xmax=127.5, ymin=127.5, ymax=-0.5] {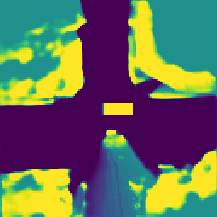};
 \nextgroupplot[
    axis background/.style={fill=white!89.8039215686275!black},
    axis line style={black},
    xmajorgrids,
    xmajorticks=false,
    xmin=-0.5, xmax=127.5,
    y dir=reverse,
    ymajorgrids,
    ymajorticks=false,
    ymin=-0.5, ymax=127.5
    ]
 \addplot graphics [includegraphics cmd=\pgfimage,xmin=-0.5, xmax=127.5, ymin=127.5, ymax=-0.5] {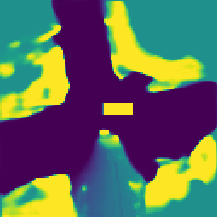};
 \nextgroupplot[
    axis background/.style={fill=white!89.8039215686275!black},
    axis line style={black},
    xmajorgrids,
    xmajorticks=false,
    xmin=-0.5, xmax=127.5,
    y dir=reverse,
    ymajorgrids,
    ymajorticks=false,
    ymin=-0.5, ymax=127.5
    ]
 \addplot graphics [includegraphics cmd=\pgfimage,xmin=-0.5, xmax=127.5, ymin=127.5, ymax=-0.5] {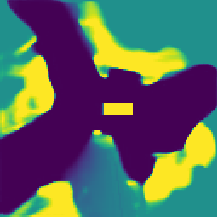};
 \nextgroupplot[
    axis background/.style={fill=white!89.8039215686275!black},
    axis line style={black},
    xmajorgrids,
    xmajorticks=false,
    xmin=-0.5, xmax=127.5,
    y dir=reverse,
    ymajorgrids,
    ymajorticks=false,
    ymin=-0.5, ymax=127.5
    ]
 \addplot graphics [includegraphics cmd=\pgfimage,xmin=-0.5, xmax=127.5, ymin=127.5, ymax=-0.5] {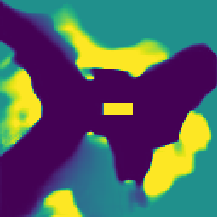};
 \nextgroupplot[
    axis background/.style={fill=white!89.8039215686275!black},
    axis line style={black},
    xmajorgrids,
    xmajorticks=false,
    xmin=-0.5, xmax=127.5,
    y dir=reverse,
    ymajorgrids,
    ymajorticks=false,
    ymin=-0.5, ymax=127.5
    ]
 \addplot graphics [includegraphics cmd=\pgfimage,xmin=-0.5, xmax=127.5, ymin=127.5, ymax=-0.5] {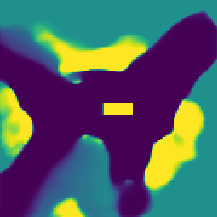};
\nextgroupplot[
    ylabel={\scalebox{.8}{SAAConvLSTM}},
    axis background/.style={fill=white!89.8039215686275!black},
    axis line style={black},
    xmajorgrids,
    xmajorticks=false,
    xmin=-0.5, xmax=127.5,
    y dir=reverse,
    ymajorgrids,
    ymajorticks=false,
    ymin=-0.5, ymax=127.5
    ]
 \addplot graphics [includegraphics cmd=\pgfimage,xmin=-0.5, xmax=127.5, ymin=127.5, ymax=-0.5] {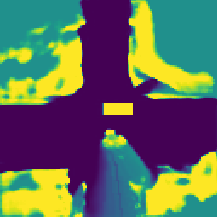};
 \nextgroupplot[
    axis background/.style={fill=white!89.8039215686275!black},
    axis line style={black},
    xmajorgrids,
    xmajorticks=false,
    xmin=-0.5, xmax=127.5,
    y dir=reverse,
    ymajorgrids,
    ymajorticks=false,
    ymin=-0.5, ymax=127.5
    ]
 \addplot graphics [includegraphics cmd=\pgfimage,xmin=-0.5, xmax=127.5, ymin=127.5, ymax=-0.5] {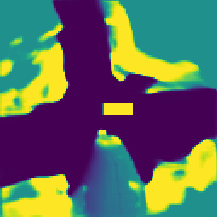};
 \nextgroupplot[
    axis background/.style={fill=white!89.8039215686275!black},
    axis line style={black},
    xmajorgrids,
    xmajorticks=false,
    xmin=-0.5, xmax=127.5,
    y dir=reverse,
    ymajorgrids,
    ymajorticks=false,
    ymin=-0.5, ymax=127.5
    ]
 \addplot graphics [includegraphics cmd=\pgfimage,xmin=-0.5, xmax=127.5, ymin=127.5, ymax=-0.5] {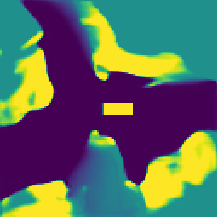};
 \nextgroupplot[
    axis background/.style={fill=white!89.8039215686275!black},
    axis line style={black},
    xmajorgrids,
    xmajorticks=false,
    xmin=-0.5, xmax=127.5,
    y dir=reverse,
    ymajorgrids,
    ymajorticks=false,
    ymin=-0.5, ymax=127.5
    ]
 \addplot graphics [includegraphics cmd=\pgfimage,xmin=-0.5, xmax=127.5, ymin=127.5, ymax=-0.5] {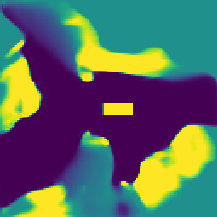};
 \nextgroupplot[
    axis background/.style={fill=white!89.8039215686275!black},
    axis line style={black},
    xmajorgrids,
    xmajorticks=false,
    xmin=-0.5, xmax=127.5,
    y dir=reverse,
    ymajorgrids,
    ymajorticks=false,
    ymin=-0.5, ymax=127.5
    ]
 \addplot graphics [includegraphics cmd=\pgfimage,xmin=-0.5, xmax=127.5, ymin=127.5, ymax=-0.5] {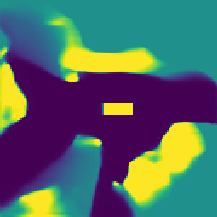};
\nextgroupplot[
    ylabel={\scalebox{.8}{PredNet}},
    axis background/.style={fill=white!89.8039215686275!black},
    axis line style={black},
    xmajorgrids,
    xmajorticks=false,
    xmin=-0.5, xmax=127.5,
    y dir=reverse,
    ymajorgrids,
    ymajorticks=false,
    ymin=-0.5, ymax=127.5
    ]
 \addplot graphics [includegraphics cmd=\pgfimage,xmin=-0.5, xmax=127.5, ymin=127.5, ymax=-0.5] {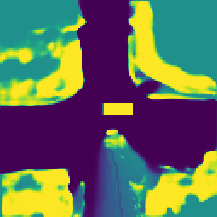};
 \nextgroupplot[
    axis background/.style={fill=white!89.8039215686275!black},
    axis line style={black},
    xmajorgrids,
    xmajorticks=false,
    xmin=-0.5, xmax=127.5,
    y dir=reverse,
    ymajorgrids,
    ymajorticks=false,
    ymin=-0.5, ymax=127.5
    ]
 \addplot graphics [includegraphics cmd=\pgfimage,xmin=-0.5, xmax=127.5, ymin=127.5, ymax=-0.5] {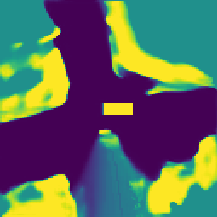};
 \nextgroupplot[
    axis background/.style={fill=white!89.8039215686275!black},
    axis line style={black},
    xmajorgrids,
    xmajorticks=false,
    xmin=-0.5, xmax=127.5,
    y dir=reverse,
    ymajorgrids,
    ymajorticks=false,
    ymin=-0.5, ymax=127.5
    ]
 \addplot graphics [includegraphics cmd=\pgfimage,xmin=-0.5, xmax=127.5, ymin=127.5, ymax=-0.5] {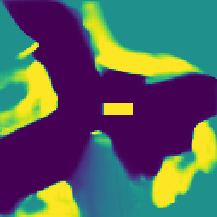};
 \nextgroupplot[
    axis background/.style={fill=white!89.8039215686275!black},
    axis line style={black},
    xmajorgrids,
    xmajorticks=false,
    xmin=-0.5, xmax=127.5,
    y dir=reverse,
    ymajorgrids,
    ymajorticks=false,
    ymin=-0.5, ymax=127.5
    ]
 \addplot graphics [includegraphics cmd=\pgfimage,xmin=-0.5, xmax=127.5, ymin=127.5, ymax=-0.5] {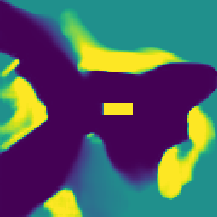};
 \nextgroupplot[
    axis background/.style={fill=white!89.8039215686275!black},
    axis line style={black},
    xmajorgrids,
    xmajorticks=false,
    xmin=-0.5, xmax=127.5,
    y dir=reverse,
    ymajorgrids,
    ymajorticks=false,
    ymin=-0.5, ymax=127.5
    ]
 \addplot graphics [includegraphics cmd=\pgfimage,xmin=-0.5, xmax=127.5, ymin=127.5, ymax=-0.5] {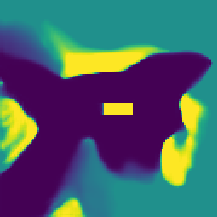};
\nextgroupplot[
    ylabel={\scalebox{.8}{PredRNN++}},
    axis background/.style={fill=white!89.8039215686275!black},
    axis line style={black},
    xmajorgrids,
    xmajorticks=false,
    xmin=-0.5, xmax=127.5,
    y dir=reverse,
    ymajorgrids,
    ymajorticks=false,
    ymin=-0.5, ymax=127.5
    ]
 \addplot graphics [includegraphics cmd=\pgfimage,xmin=-0.5, xmax=127.5, ymin=127.5, ymax=-0.5] {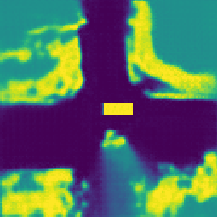};
 
 \nextgroupplot[
    axis background/.style={fill=white!89.8039215686275!black},
    axis line style={black},
    xmajorgrids,
    xmajorticks=false,
    xmin=-0.5, xmax=127.5,
    y dir=reverse,
    ymajorgrids,
    ymajorticks=false,
    ymin=-0.5, ymax=127.5
    ]
 \addplot graphics [includegraphics cmd=\pgfimage,xmin=-0.5, xmax=127.5, ymin=127.5, ymax=-0.5] {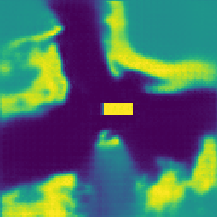};
 
 \nextgroupplot[
    axis background/.style={fill=white!89.8039215686275!black},
    axis line style={black},
    xmajorgrids,
    xmajorticks=false,
    xmin=-0.5, xmax=127.5,
    y dir=reverse,
    ymajorgrids,
    ymajorticks=false,
    ymin=-0.5, ymax=127.5
    ]
 \addplot graphics [includegraphics cmd=\pgfimage,xmin=-0.5, xmax=127.5, ymin=127.5, ymax=-0.5] {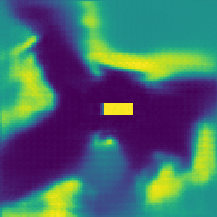};
 
 \nextgroupplot[
    axis background/.style={fill=white!89.8039215686275!black},
    axis line style={black},
    xmajorgrids,
    xmajorticks=false,
    xmin=-0.5, xmax=127.5,
    y dir=reverse,
    ymajorgrids,
    ymajorticks=false,
    ymin=-0.5, ymax=127.5
    ]
 \addplot graphics [includegraphics cmd=\pgfimage,xmin=-0.5, xmax=127.5, ymin=127.5, ymax=-0.5] {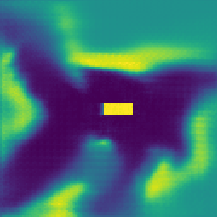};
 
 \nextgroupplot[
    axis background/.style={fill=white!89.8039215686275!black},
    axis line style={black},
    xmajorgrids,
    xmajorticks=false,
    xmin=-0.5, xmax=127.5,
    y dir=reverse,
    ymajorgrids,
    ymajorticks=false,
    ymin=-0.5, ymax=127.5
    ]
 \addplot graphics [includegraphics cmd=\pgfimage,xmin=-0.5, xmax=127.5, ymin=127.5, ymax=-0.5] {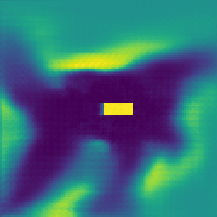};
 
  \nextgroupplot[
    ylabel={\scalebox{.8}{CDNA}},
    axis background/.style={fill=white!89.8039215686275!black},
    axis line style={black},
    xmajorgrids,
    xmajorticks=false,
    xmin=-0.5, xmax=127.5,
    y dir=reverse,
    ymajorgrids,
    ymajorticks=false,
    ymin=-0.5, ymax=127.5
    ]
 \addplot graphics [includegraphics cmd=\pgfimage,xmin=-0.5, xmax=127.5, ymin=127.5, ymax=-0.5] {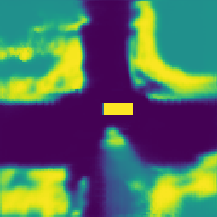};
 
  \nextgroupplot[
    axis background/.style={fill=white!89.8039215686275!black},
    axis line style={black},
    xmajorgrids,
    xmajorticks=false,
    xmin=-0.5, xmax=127.5,
    y dir=reverse,
    ymajorgrids,
    ymajorticks=false,
    ymin=-0.5, ymax=127.5
    ]
 \addplot graphics [includegraphics cmd=\pgfimage,xmin=-0.5, xmax=127.5, ymin=127.5, ymax=-0.5] {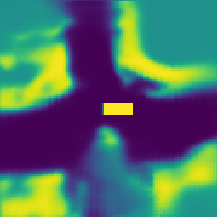};
 
  \nextgroupplot[
    axis background/.style={fill=white!89.8039215686275!black},
    axis line style={black},
    xmajorgrids,
    xmajorticks=false,
    xmin=-0.5, xmax=127.5,
    y dir=reverse,
    ymajorgrids,
    ymajorticks=false,
    ymin=-0.5, ymax=127.5
    ]
 \addplot graphics [includegraphics cmd=\pgfimage,xmin=-0.5, xmax=127.5, ymin=127.5, ymax=-0.5] {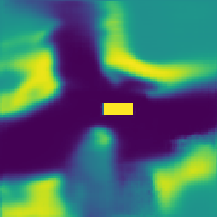};
 
  \nextgroupplot[
    axis background/.style={fill=white!89.8039215686275!black},
    axis line style={black},
    xmajorgrids,
    xmajorticks=false,
    xmin=-0.5, xmax=127.5,
    y dir=reverse,
    ymajorgrids,
    ymajorticks=false,
    ymin=-0.5, ymax=127.5
    ]
 \addplot graphics [includegraphics cmd=\pgfimage,xmin=-0.5, xmax=127.5, ymin=127.5, ymax=-0.5] {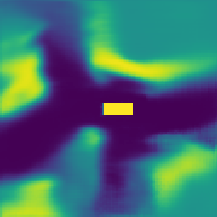};
 
  \nextgroupplot[
    axis background/.style={fill=white!89.8039215686275!black},
    axis line style={black},
    xmajorgrids,
    xmajorticks=false,
    xmin=-0.5, xmax=127.5,
    y dir=reverse,
    ymajorgrids,
    ymajorticks=false,
    ymin=-0.5, ymax=127.5
    ]
 \addplot graphics [includegraphics cmd=\pgfimage,xmin=-0.5, xmax=127.5, ymin=127.5, ymax=-0.5] {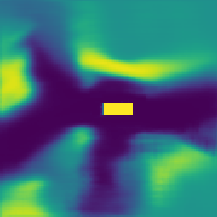};
\end{groupplot}

\begin{groupplot}[
  group style={
      group name=right plots,
      group size=5 by 1,
      horizontal sep=0pt,
      x descriptions at=edge bottom},
  width=1.25cm,
  height=1.25cm,
  scale only axis,
  every axis title/.style={yshift=4pt, xshift=17.7pt}]
   
\nextgroupplot[
anchor=north west, at={($(left plots c5r1.north east) + (0.5cm,0.0cm)$)},
title={$t=-4$},
axis background/.style={fill=white!89.8039215686275!black},
axis line style={black},
xmajorgrids,
xmajorticks=false,
xmin=-0.5, xmax=127.5,
y dir=reverse,
ymajorgrids,
ymajorticks=false,
ymin=-0.5, ymax=127.5
]
\addplot graphics [includegraphics cmd=\pgfimage,xmin=-0.5, xmax=127.5, ymin=127.5, ymax=-0.5] {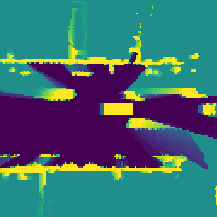};

\nextgroupplot[
title={$t=-3$},
axis background/.style={fill=white!89.8039215686275!black},
axis line style={black},
xmajorgrids,
xmajorticks=false,
xmin=-0.5, xmax=127.5,
y dir=reverse,
ymajorgrids,
ymajorticks=false,
ymin=-0.5, ymax=127.5
]
\addplot graphics [includegraphics cmd=\pgfimage,xmin=-0.5, xmax=127.5, ymin=127.5, ymax=-0.5] {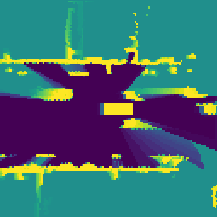};

\nextgroupplot[
title={$t=-2$},
axis background/.style={fill=white!89.8039215686275!black},
axis line style={black},
xmajorgrids,
xmajorticks=false,
xmin=-0.5, xmax=127.5,
y dir=reverse,
ymajorgrids,
ymajorticks=false,
ymin=-0.5, ymax=127.5
]
\addplot graphics [includegraphics cmd=\pgfimage,xmin=-0.5, xmax=127.5, ymin=127.5, ymax=-0.5] {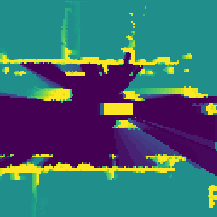};

\nextgroupplot[
title={$t=-1$},
axis background/.style={fill=white!89.8039215686275!black},
axis line style={black},
xmajorgrids,
xmajorticks=false,
xmin=-0.5, xmax=127.5,
y dir=reverse,
ymajorgrids,
ymajorticks=false,
ymin=-0.5, ymax=127.5
]
\addplot graphics [includegraphics cmd=\pgfimage,xmin=-0.5, xmax=127.5, ymin=127.5, ymax=-0.5] {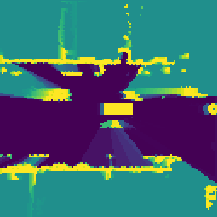};

\nextgroupplot[
title={$t=0$},
axis background/.style={fill=white!89.8039215686275!black},
axis line style={black},
xmajorgrids,
xmajorticks=false,
xmin=-0.5, xmax=127.5,
y dir=reverse,
ymajorgrids,
ymajorticks=false,
ymin=-0.5, ymax=127.5
]
\addplot graphics [includegraphics cmd=\pgfimage,xmin=-0.5, xmax=127.5, ymin=127.5, ymax=-0.5] {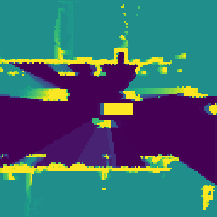};

\end{groupplot}

\begin{groupplot}[
    group style={
      group size=5 by 6,
      horizontal sep=0pt,
      vertical sep = 0pt,
      x descriptions at=edge bottom},
    width=1.25cm,
    height=1.25cm,
    scale only axis,
    ytick pos=left,
    every axis title/.style={yshift=4pt, xshift=17.7pt}]
 \nextgroupplot[
    anchor=north west, at={($(right plots c1r1.south west) - (0.0cm,0.6cm)$)},
    title={$t=5$},
    axis background/.style={fill=white!89.8039215686275!black},
    axis line style={black},
    xmajorgrids,
    xmajorticks=false,
    xmin=-0.5, xmax=127.5,
    y dir=reverse,
    ymajorgrids,
    ymajorticks=false,
    ymin=-0.5, ymax=127.5
    ]
  \addplot graphics [includegraphics cmd=\pgfimage,xmin=-0.5, xmax=127.5, ymin=127.5, ymax=-0.5] {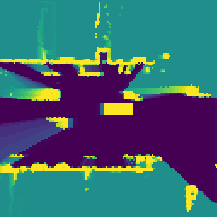};
 
  \nextgroupplot[
    title={$t=10$},
    axis background/.style={fill=white!89.8039215686275!black},
    axis line style={black},
    xmajorgrids,
    xmajorticks=false,
    xmin=-0.5, xmax=127.5,
    y dir=reverse,
    ymajorgrids,
    ymajorticks=false,
    ymin=-0.5, ymax=127.5
    ]
 \addplot graphics [includegraphics cmd=\pgfimage,xmin=-0.5, xmax=127.5, ymin=127.5, ymax=-0.5] {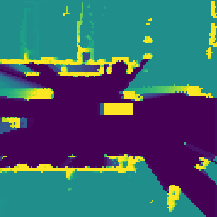};
 
 \nextgroupplot[
    title={$t=15$},
    axis background/.style={fill=white!89.8039215686275!black},
    axis line style={black},
    xmajorgrids,
    xmajorticks=false,
    xmin=-0.5, xmax=127.5,
    y dir=reverse,
    ymajorgrids,
    ymajorticks=false,
    ymin=-0.5, ymax=127.5
    ]
 \addplot graphics [includegraphics cmd=\pgfimage,xmin=-0.5, xmax=127.5, ymin=127.5, ymax=-0.5] {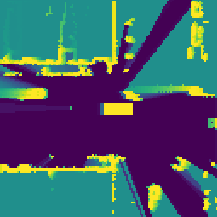};
 \nextgroupplot[
    title={$t=20$},
    axis background/.style={fill=white!89.8039215686275!black},
    axis line style={black},
    xmajorgrids,
    xmajorticks=false,
    xmin=-0.5, xmax=127.5,
    y dir=reverse,
    ymajorgrids,
    ymajorticks=false,
    ymin=-0.5, ymax=127.5
    ]
 \addplot graphics [includegraphics cmd=\pgfimage,xmin=-0.5, xmax=127.5, ymin=127.5, ymax=-0.5] {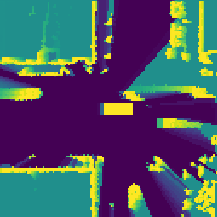};
 \nextgroupplot[
    title={$t=25$},
    axis background/.style={fill=white!89.8039215686275!black},
    axis line style={black},
    xmajorgrids,
    xmajorticks=false,
    xmin=-0.5, xmax=127.5,
    y dir=reverse,
    ymajorgrids,
    ymajorticks=false,
    ymin=-0.5, ymax=127.5
    ]
 \addplot graphics [includegraphics cmd=\pgfimage,xmin=-0.5, xmax=127.5, ymin=127.5, ymax=-0.5] {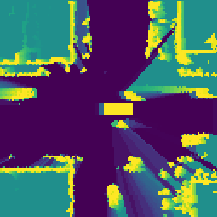};
\nextgroupplot[
    axis background/.style={fill=white!89.8039215686275!black},
    axis line style={black},
    xmajorgrids,
    xmajorticks=false,
    xmin=-0.5, xmax=127.5,
    y dir=reverse,
    ymajorgrids,
    ymajorticks=false,
    ymin=-0.5, ymax=127.5
    ]
 \addplot graphics [includegraphics cmd=\pgfimage,xmin=-0.5, xmax=127.5, ymin=127.5, ymax=-0.5] {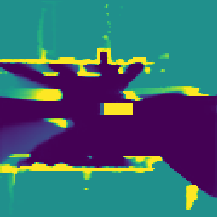};
 \nextgroupplot[
    axis background/.style={fill=white!89.8039215686275!black},
    axis line style={black},
    xmajorgrids,
    xmajorticks=false,
    xmin=-0.5, xmax=127.5,
    y dir=reverse,
    ymajorgrids,
    ymajorticks=false,
    ymin=-0.5, ymax=127.5
    ]
 \addplot graphics [includegraphics cmd=\pgfimage,xmin=-0.5, xmax=127.5, ymin=127.5, ymax=-0.5] {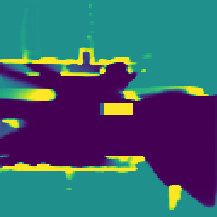};
 \nextgroupplot[
    axis background/.style={fill=white!89.8039215686275!black},
    axis line style={black},
    xmajorgrids,
    xmajorticks=false,
    xmin=-0.5, xmax=127.5,
    y dir=reverse,
    ymajorgrids,
    ymajorticks=false,
    ymin=-0.5, ymax=127.5
    ]
 \addplot graphics [includegraphics cmd=\pgfimage,xmin=-0.5, xmax=127.5, ymin=127.5, ymax=-0.5] {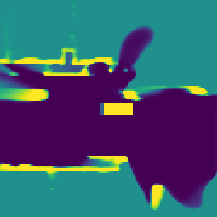};
 \nextgroupplot[
    axis background/.style={fill=white!89.8039215686275!black},
    axis line style={black},
    xmajorgrids,
    xmajorticks=false,
    xmin=-0.5, xmax=127.5,
    y dir=reverse,
    ymajorgrids,
    ymajorticks=false,
    ymin=-0.5, ymax=127.5
    ]
 \addplot graphics [includegraphics cmd=\pgfimage,xmin=-0.5, xmax=127.5, ymin=127.5, ymax=-0.5] {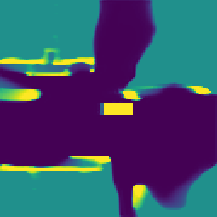};
 \nextgroupplot[
    axis background/.style={fill=white!89.8039215686275!black},
    axis line style={black},
    xmajorgrids,
    xmajorticks=false,
    xmin=-0.5, xmax=127.5,
    y dir=reverse,
    ymajorgrids,
    ymajorticks=false,
    ymin=-0.5, ymax=127.5
    ]
 \addplot graphics [includegraphics cmd=\pgfimage,xmin=-0.5, xmax=127.5, ymin=127.5, ymax=-0.5] {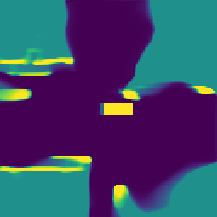};
\nextgroupplot[
    axis background/.style={fill=white!89.8039215686275!black},
    axis line style={black},
    xmajorgrids,
    xmajorticks=false,
    xmin=-0.5, xmax=127.5,
    y dir=reverse,
    ymajorgrids,
    ymajorticks=false,
    ymin=-0.5, ymax=127.5
    ]
 \addplot graphics [includegraphics cmd=\pgfimage,xmin=-0.5, xmax=127.5, ymin=127.5, ymax=-0.5] {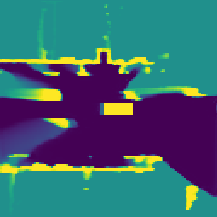};
 \nextgroupplot[
    axis background/.style={fill=white!89.8039215686275!black},
    axis line style={black},
    xmajorgrids,
    xmajorticks=false,
    xmin=-0.5, xmax=127.5,
    y dir=reverse,
    ymajorgrids,
    ymajorticks=false,
    ymin=-0.5, ymax=127.5
    ]
 \addplot graphics [includegraphics cmd=\pgfimage,xmin=-0.5, xmax=127.5, ymin=127.5, ymax=-0.5] {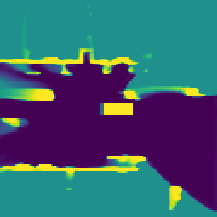};
 \nextgroupplot[
    axis background/.style={fill=white!89.8039215686275!black},
    axis line style={black},
    xmajorgrids,
    xmajorticks=false,
    xmin=-0.5, xmax=127.5,
    y dir=reverse,
    ymajorgrids,
    ymajorticks=false,
    ymin=-0.5, ymax=127.5
    ]
 \addplot graphics [includegraphics cmd=\pgfimage,xmin=-0.5, xmax=127.5, ymin=127.5, ymax=-0.5] {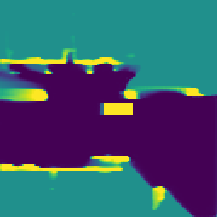};
 \nextgroupplot[
    axis background/.style={fill=white!89.8039215686275!black},
    axis line style={black},
    xmajorgrids,
    xmajorticks=false,
    xmin=-0.5, xmax=127.5,
    y dir=reverse,
    ymajorgrids,
    ymajorticks=false,
    ymin=-0.5, ymax=127.5
    ]
 \addplot graphics [includegraphics cmd=\pgfimage,xmin=-0.5, xmax=127.5, ymin=127.5, ymax=-0.5] {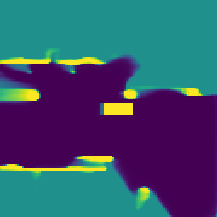};
 \nextgroupplot[
    axis background/.style={fill=white!89.8039215686275!black},
    axis line style={black},
    xmajorgrids,
    xmajorticks=false,
    xmin=-0.5, xmax=127.5,
    y dir=reverse,
    ymajorgrids,
    ymajorticks=false,
    ymin=-0.5, ymax=127.5
    ]
 \addplot graphics [includegraphics cmd=\pgfimage,xmin=-0.5, xmax=127.5, ymin=127.5, ymax=-0.5] {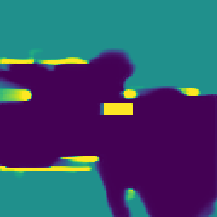};
\nextgroupplot[
    axis background/.style={fill=white!89.8039215686275!black},
    axis line style={black},
    xmajorgrids,
    xmajorticks=false,
    xmin=-0.5, xmax=127.5,
    y dir=reverse,
    ymajorgrids,
    ymajorticks=false,
    ymin=-0.5, ymax=127.5
    ]
 \addplot graphics [includegraphics cmd=\pgfimage,xmin=-0.5, xmax=127.5, ymin=127.5, ymax=-0.5] {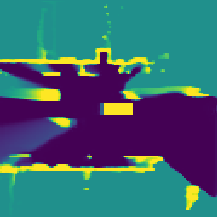};
 \nextgroupplot[
    axis background/.style={fill=white!89.8039215686275!black},
    axis line style={black},
    xmajorgrids,
    xmajorticks=false,
    xmin=-0.5, xmax=127.5,
    y dir=reverse,
    ymajorgrids,
    ymajorticks=false,
    ymin=-0.5, ymax=127.5
    ]
 \addplot graphics [includegraphics cmd=\pgfimage,xmin=-0.5, xmax=127.5, ymin=127.5, ymax=-0.5] {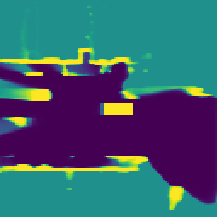};
 \nextgroupplot[
    axis background/.style={fill=white!89.8039215686275!black},
    axis line style={black},
    xmajorgrids,
    xmajorticks=false,
    xmin=-0.5, xmax=127.5,
    y dir=reverse,
    ymajorgrids,
    ymajorticks=false,
    ymin=-0.5, ymax=127.5
    ]
 \addplot graphics [includegraphics cmd=\pgfimage,xmin=-0.5, xmax=127.5, ymin=127.5, ymax=-0.5] {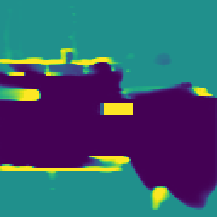};
 \nextgroupplot[
    axis background/.style={fill=white!89.8039215686275!black},
    axis line style={black},
    xmajorgrids,
    xmajorticks=false,
    xmin=-0.5, xmax=127.5,
    y dir=reverse,
    ymajorgrids,
    ymajorticks=false,
    ymin=-0.5, ymax=127.5
    ]
 \addplot graphics [includegraphics cmd=\pgfimage,xmin=-0.5, xmax=127.5, ymin=127.5, ymax=-0.5] {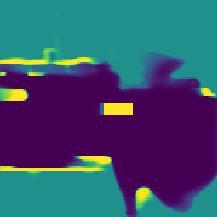};
 \nextgroupplot[
    axis background/.style={fill=white!89.8039215686275!black},
    axis line style={black},
    xmajorgrids,
    xmajorticks=false,
    xmin=-0.5, xmax=127.5,
    y dir=reverse,
    ymajorgrids,
    ymajorticks=false,
    ymin=-0.5, ymax=127.5
    ]
 \addplot graphics [includegraphics cmd=\pgfimage,xmin=-0.5, xmax=127.5, ymin=127.5, ymax=-0.5] {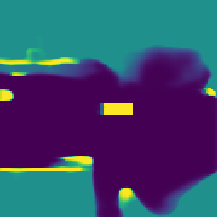};
\nextgroupplot[
    axis background/.style={fill=white!89.8039215686275!black},
    axis line style={black},
    xmajorgrids,
    xmajorticks=false,
    xmin=-0.5, xmax=127.5,
    y dir=reverse,
    ymajorgrids,
    ymajorticks=false,
    ymin=-0.5, ymax=127.5
    ]
 \addplot graphics [includegraphics cmd=\pgfimage,xmin=-0.5, xmax=127.5, ymin=127.5, ymax=-0.5] {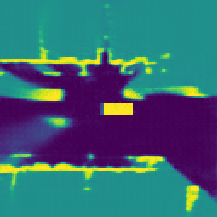};
 
 \nextgroupplot[
    axis background/.style={fill=white!89.8039215686275!black},
    axis line style={black},
    xmajorgrids,
    xmajorticks=false,
    xmin=-0.5, xmax=127.5,
    y dir=reverse,
    ymajorgrids,
    ymajorticks=false,
    ymin=-0.5, ymax=127.5
    ]
 \addplot graphics [includegraphics cmd=\pgfimage,xmin=-0.5, xmax=127.5, ymin=127.5, ymax=-0.5] {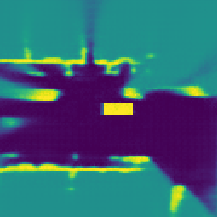};
 
 \nextgroupplot[
    axis background/.style={fill=white!89.8039215686275!black},
    axis line style={black},
    xmajorgrids,
    xmajorticks=false,
    xmin=-0.5, xmax=127.5,
    y dir=reverse,
    ymajorgrids,
    ymajorticks=false,
    ymin=-0.5, ymax=127.5
    ]
 \addplot graphics [includegraphics cmd=\pgfimage,xmin=-0.5, xmax=127.5, ymin=127.5, ymax=-0.5] {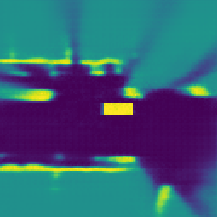};
 
 \nextgroupplot[
    axis background/.style={fill=white!89.8039215686275!black},
    axis line style={black},
    xmajorgrids,
    xmajorticks=false,
    xmin=-0.5, xmax=127.5,
    y dir=reverse,
    ymajorgrids,
    ymajorticks=false,
    ymin=-0.5, ymax=127.5
    ]
 \addplot graphics [includegraphics cmd=\pgfimage,xmin=-0.5, xmax=127.5, ymin=127.5, ymax=-0.5] {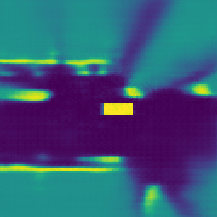};
 
 \nextgroupplot[
    axis background/.style={fill=white!89.8039215686275!black},
    axis line style={black},
    xmajorgrids,
    xmajorticks=false,
    xmin=-0.5, xmax=127.5,
    y dir=reverse,
    ymajorgrids,
    ymajorticks=false,
    ymin=-0.5, ymax=127.5
    ]
 \addplot graphics [includegraphics cmd=\pgfimage,xmin=-0.5, xmax=127.5, ymin=127.5, ymax=-0.5] {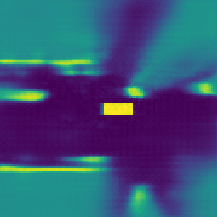};
 
  \nextgroupplot[
    axis background/.style={fill=white!89.8039215686275!black},
    axis line style={black},
    xmajorgrids,
    xmajorticks=false,
    xmin=-0.5, xmax=127.5,
    y dir=reverse,
    ymajorgrids,
    ymajorticks=false,
    ymin=-0.5, ymax=127.5
    ]
 \addplot graphics [includegraphics cmd=\pgfimage,xmin=-0.5, xmax=127.5, ymin=127.5, ymax=-0.5] {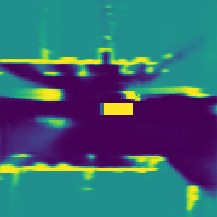};
 
  \nextgroupplot[
    axis background/.style={fill=white!89.8039215686275!black},
    axis line style={black},
    xmajorgrids,
    xmajorticks=false,
    xmin=-0.5, xmax=127.5,
    y dir=reverse,
    ymajorgrids,
    ymajorticks=false,
    ymin=-0.5, ymax=127.5
    ]
 \addplot graphics [includegraphics cmd=\pgfimage,xmin=-0.5, xmax=127.5, ymin=127.5, ymax=-0.5] {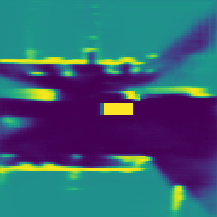};
 
  \nextgroupplot[
    axis background/.style={fill=white!89.8039215686275!black},
    axis line style={black},
    xmajorgrids,
    xmajorticks=false,
    xmin=-0.5, xmax=127.5,
    y dir=reverse,
    ymajorgrids,
    ymajorticks=false,
    ymin=-0.5, ymax=127.5
    ]
 \addplot graphics [includegraphics cmd=\pgfimage,xmin=-0.5, xmax=127.5, ymin=127.5, ymax=-0.5] {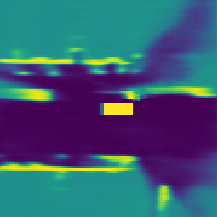};
 
  \nextgroupplot[
    axis background/.style={fill=white!89.8039215686275!black},
    axis line style={black},
    xmajorgrids,
    xmajorticks=false,
    xmin=-0.5, xmax=127.5,
    y dir=reverse,
    ymajorgrids,
    ymajorticks=false,
    ymin=-0.5, ymax=127.5
    ]
 \addplot graphics [includegraphics cmd=\pgfimage,xmin=-0.5, xmax=127.5, ymin=127.5, ymax=-0.5] {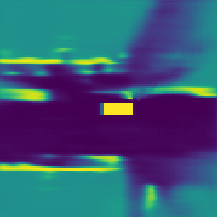};
 
  \nextgroupplot[
    axis background/.style={fill=white!89.8039215686275!black},
    axis line style={black},
    xmajorgrids,
    xmajorticks=false,
    xmin=-0.5, xmax=127.5,
    y dir=reverse,
    ymajorgrids,
    ymajorticks=false,
    ymin=-0.5, ymax=127.5
    ]
 \addplot graphics [includegraphics cmd=\pgfimage,xmin=-0.5, xmax=127.5, ymin=127.5, ymax=-0.5] {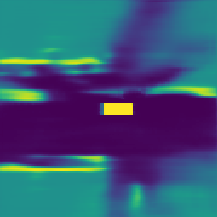};

\end{groupplot}
\begin{groupplot}[
    group style={
      group name=legend,
      group size=1 by 1,
      horizontal sep=0pt,
      vertical sep = 0pt,
      x descriptions at=edge bottom},
    width=5.23cm,
    height=0.5cm,
    scale only axis,
    ytick pos=left,
    every axis title/.style={yshift=4pt, xshift=17.7pt}]
 \nextgroupplot[
    anchor=north west, at={($(left bottom c5r6.south west) - (1cm,0.25cm)$)},
    axis background/.style={fill=white!89.8039215686275!black},
    axis line style={white},
    xmajorgrids,
    xmajorticks=false,
    xmin=-0.5, xmax=127.5,
    y dir=reverse,
    ymajorgrids,
    ymajorticks=false,
    ymin=-0.5, ymax=127.5
    ]
  \addplot graphics [includegraphics cmd=\pgfimage,xmin=-0.5, xmax=127.5, ymin=127.5, ymax=-0.5] {figures/legend.png};
 \end{groupplot}
\end{tikzpicture}}
\caption{Example \SI{2.5}{\second} predictions on the Waymo dataset and the associated ground truth. Left: The ego-vehicle makes a turn at an intersection. Right: The ego-vehicle approaches an intersection. In both cases, our approaches maintain the presence of the other vehicle at the intersection, in contrast to the baseline predictions. TAAConvLSTM also better captures the static environment rotation and translation.}
\label{fig:WaymoVis}
\end{center}
\end{figure*}

%% file: sections/06-discussion.tex
\section{Discussion and Future Work}
\label{sec:discussion}
Our experiments show that the addition of the TAAConv and SAAConv mechanisms to the ConvLSTM architecture reduces the vanishing of moving objects and increases the overall prediction quality of the model. TAAConv enriches the hidden representation of the ConvLSTM and enables correct motion prediction for moving objects while maintaining their original shape (\cref{fig:KittiVis} Left). TAAConvLSTM also excels at the relative static environment rotation and inference. It correctly captures the intersection rotation during the ego-vehicle turn (\cref{fig:WaymoVis} Left) and infers that there is an incoming intersection (\cref{fig:WaymoVis} Right). We hypothesize that the mechanism either encountered similar intersections during training, or learned to infer the semantic layout of the map based on the motion of the other traffic participants.
SAAConvLSTM maintains higher overall prediction quality with a smaller number of parameters than all other considered approaches. It models the ego-vehicle motion within the environment the most accurately by correctly following the layout of the road (\cref{fig:KittiVis} Right).

Both experiments in \cref{sec:kitti} and \cref{sec:waymo} highlight the challenges of evaluating environment prediction. There is a trade-off between the IS and MSE metrics, which is often present in prediction tasks due to the multimodal characteristics of the problem. The ConvLSTM used in vanilla PredNet tends to blur and remove moving objects from the predictions, leading to the lowest MSE for a task with multiple possible futures. This is also visible in the PredRNN++ predictions, which strategically blur out the dynamic parts of the scene to reduce the MSE. Maintaining the dynamic objects in the prediction often results in better IS and worse MSE.
The IS metric more effectively captures the necessary properties of a valid environment prediction (e.g. object presence, scene structure) than the MSE metric, reflecting the observed qualitative performance. 

The TAAConv and SAAConv operators facilitate reasoning over multiple representational subspaces and ultimately lead to more detailed predictions with reduced object vanishing. Both our proposed methods achieve improved IS values without deteriorating the MSE scores. 
The IS performance gap between our TAAConvLSTM architecture and the baselines increases with longer time horizon predictions, demonstrating that TAAConvLSTM maintains longer-term temporal dependencies and reduces object vanishing.
Reduced vanishing of moving objects in the KITTI dataset is further supported by the MOBBM results. Our models maintain more accurate moving objects' position and shape leading to better MOBBM values than the baselines.

Even though the application of attention in this work is motivated by dynamic object disappearance, our proposed approach appears to learn to \textit{attend} to the static parts of the environment as well as the dynamic objects. The Waymo Open Dataset consists of a considerably larger number of scenarios, the majority of which contain agents with smaller relative velocities with the respect to the ego-vehicle than those in the KITTI dataset. This data imbalance results in a reduced 
quantitative improvement on the Waymo dataset, but our proposed mechanisms continue to demonstrate qualitatively superior predictions, particularly in the static environment prediction.

There are several further avenues that could be explored for future work. The occupancy grid representation can be augmented with semantic map priors. 
We can infuse prior information into the attention operator directly to enforce its interpretability. Additionally, positional-encodings for the attention can be tailored to the problem specific task, e.g. exploiting the Inertial Measurement Unit (IMU) in the autonomous driving setting.

%% file: sections/07-appendix.tex
\appendix
\label{sec:appendix}

\subsection{Example Visualization of Moving Object Vanishing in the Baseline Framework}
\label{sec:vanishing}
\begin{figure}[htb]
  \centering
  \centerline{\includegraphics[width=10cm]{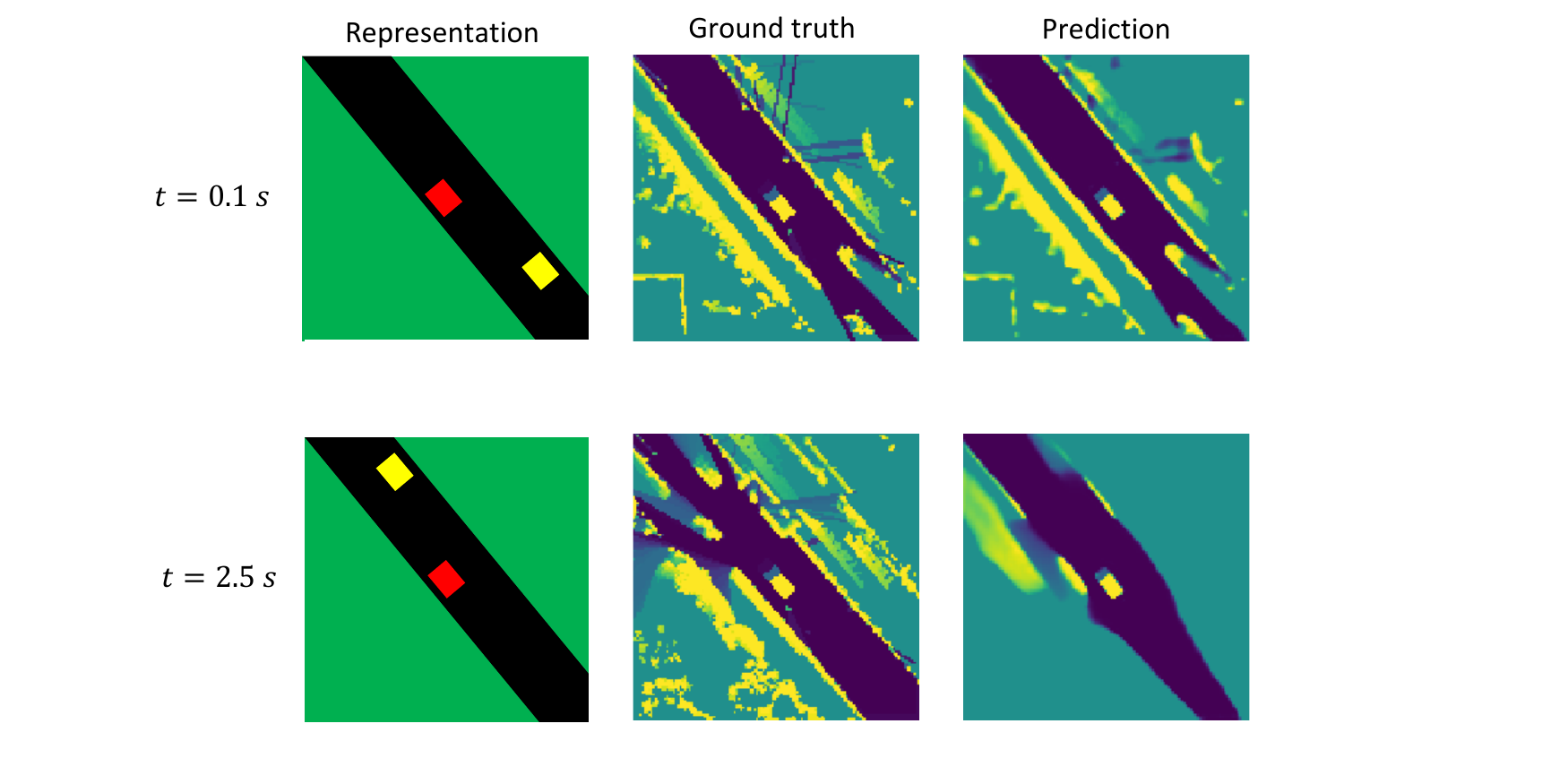}}
  \caption{Moving object vanishing. From the left: schematic world representation for illustration, ground truth occupancy grid, and predicted occupancy grid (baseline PredNet). Top row: prediction \SI{0.1}{\second} in to the future. Bottom row: \SI{2.5}{\second} in to the future. Labels used in the world representation: ego-vehicle (red), other moving vehicle (yellow), road (black), and static environment (green).}
  \label{fig:vanish}
\end{figure}

\cref{fig:vanish} shows an example of a moving object vanishing in the prediction of the PredNet~\cite{lotter2016deep} baseline. The gradual vanishing results in the complete disappearance of the moving object (yellow vehicle in \cref{fig:vanish}).

\subsection{Occupancy Grid Generation}
\label{sec:occupancy}
This section describes the procedure for generating the occupancy grid to form the input and target data for the proposed network. The occupancy grid is generated according to the procedure used by ~\citet{itkina2019dynamic}. To perform the occupancy grid update, two types of methods can be considered: Bayesian methods and Dempster-Shafer Theory (DST)~\cite{nuss2018random}. DST is a decision-making strategy that provides evidential updates in support of or against a set of the hypotheses given the evidence~\cite{gordon1984dempster}. The additional information provided by DST methods enriches our understanding of the surroundings and improves the environment prediction models~\cite{itkina2019dynamic}.


Our occupancy grid is updated following the Dempster-Shafer Theory (DST) approach described by~\citet{nuss2018random} which considers the set of possible \textit{events} defined in a \textit{frame of discernment} $\Omega$. For the occupancy grid scenario, the events are free space $F$ and occupied space $O$, i.e. $\Omega = \{F,O\}$. The possible set of hypotheses is defined as the power set of $\Omega$ which is $2^\Omega = \{\emptyset, \{F\},\{O\},\{F,O\}\}$, where $\emptyset$ denotes the empty set. Each hypothesis has a corresponding belief mass. The sum of belief masses over the hypothesis set is equal to one. The proposition $\emptyset$ can be pruned from our set because a cell cannot be neither occupied and unoccupied. The initial condition for the belief masses assigns one to the $\{F,O\}$ set representing the unknown state of the cell. Therefore, the mass over all other hypotheses is zero because there is no evidence stating that the cell is occupied or free until sensor measurements are received.

Subsequently, the DST update rule defined in \cref{update} is used with every new measurement to update the previous mass $m^{c}_{k-1}$ in cell $c$ at time step $k$ with new measurement $m^{c}_{k,z}$ \cite{nuss2018random}:
\begin{equation}
\label{update}
\begin{split}
m^{c}_k(A) = &\, m^{c}_{k-1} \oplus m^{c}_{k,z}(A) \\
     := & \frac{\sum_{X\cap Y=A} m^{c}_{k-1}(X)m^{c}_{k,z}(Y)}{1-\sum_{X\cap Y=\emptyset} m^{c}_{k-1}(X)m^{c}_{k,z}(Y)} \\
    & \forall A,X,Y\in \{\{F\},\{O\},\{F,O\}\}.  
\end{split}
\end{equation}
Before each measurement, we perform information aging to the prior masses:
\begin{equation}
\begin{split}
m^{c}_{k,\alpha}(\{O\}) & = \min(\alpha \cdot m^{c}_{k}(\{O\}),1) \\
m^{c}_{k,\alpha}(\{F\}) & = \min(\alpha \cdot m^{c}_{k}(\{F\}),1) \\
m^{c}_{k,\alpha}(\{F,O\}) & = 1 - m^{c}_{k,\alpha}(\{O\}) - m^{c}_{k,\alpha}(\{F\}).
\end{split}{}
\end{equation}
An estimate of conventional occupancy grid probabilities are acquired using the concept of pignistic probability:
\begin{equation}
\label{probability}
betP(B)=\sum_{A \in 2^\Omega} m(A)\cdot \frac{|B\cap A|}{|A|}
\end{equation}

\subsection{Architectures}
\label{sec:architectures}
In the following section, we outline the architectures considered in this paper, which are PredNet~\cite{lotter2016deep}, PredNet with TAAConvLSTM (ours), PredNet with SAAConvLSTM (ours), and PredRNN++~\cite{wang2018predrnn++}. All architectures follow a sequence-to-sequence model where we provide a number of past frames and predict the future frames. For each architecture, we provide a short overview of the framework and its hyperparameters.

\subsubsection{Predictive Coding Network (PredNet)} 
\begin{figure}
\begin{minipage}{0.45\textwidth}
    \centering
    \captionsetup{width=.9\textwidth}
    \includegraphics[width=0.7\linewidth]{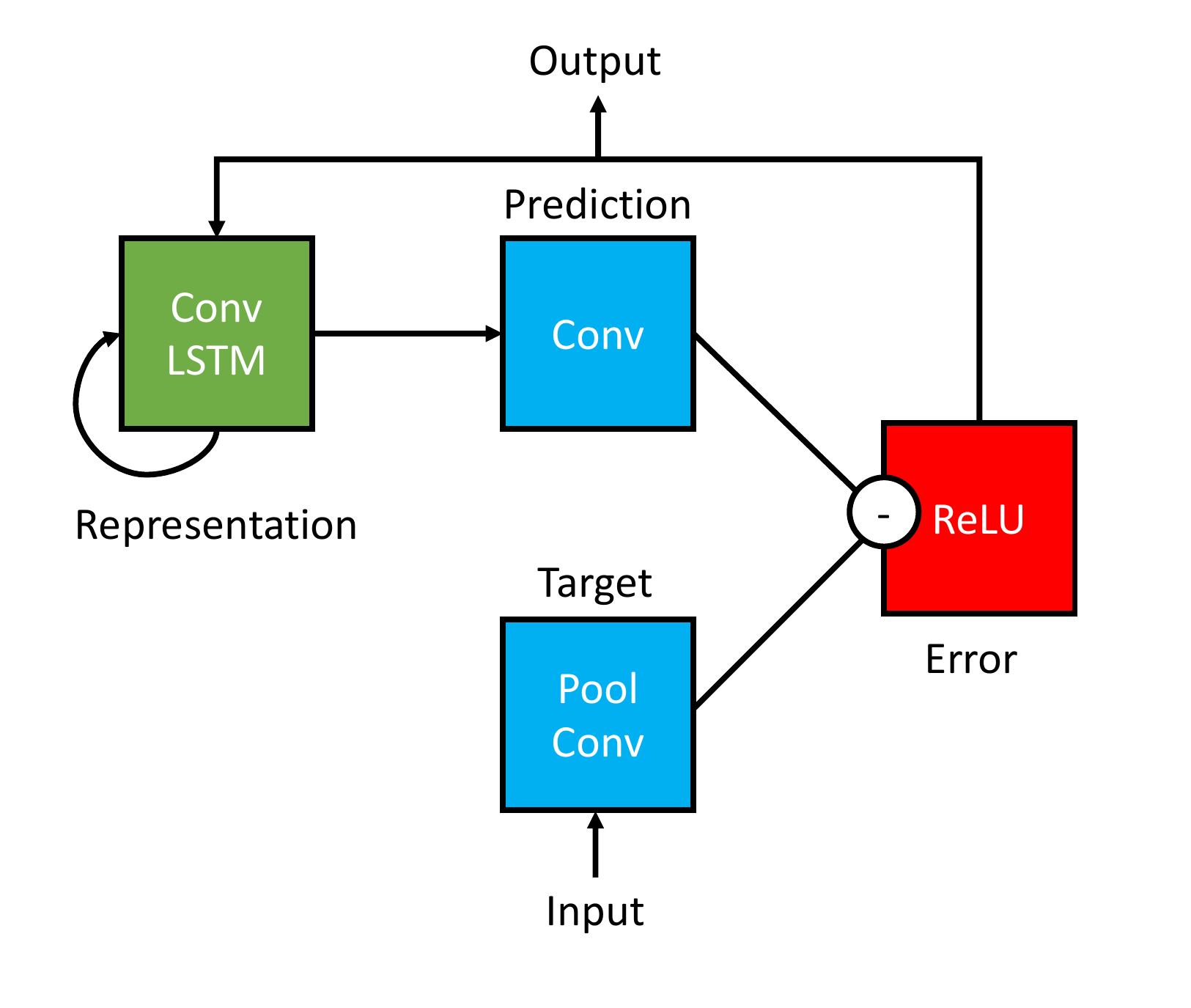}
   \caption{Prediction cell for the Predictive Coding Network (PredNet) created by \citet{lotter2016deep}.}
  \label{fig:PrednetCell}
\end{minipage}
\begin{minipage}{0.45\textwidth}
    \centering
    \captionsetup{width=.85\textwidth}
    \includegraphics[width=0.7\linewidth]{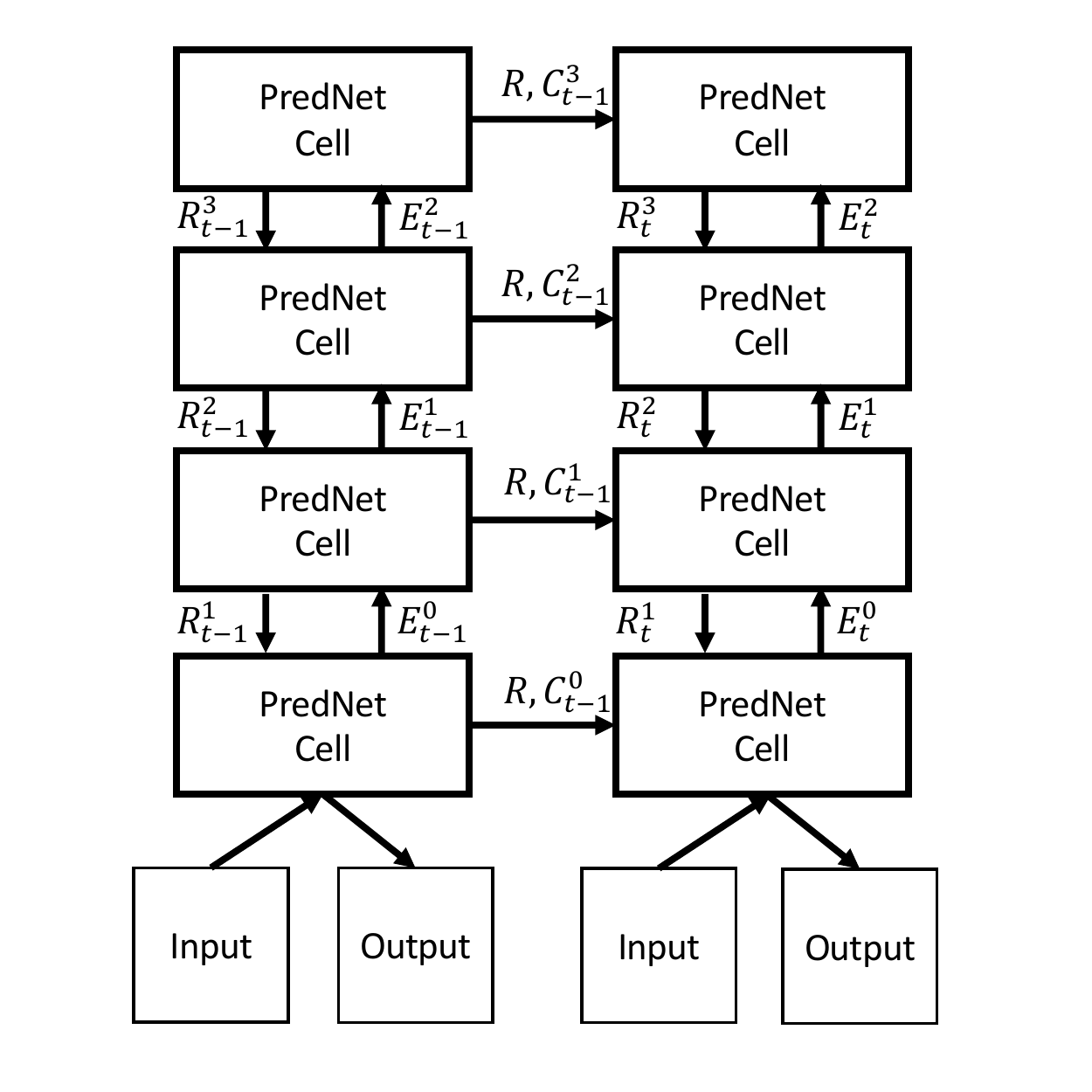}
   \caption{The architecture with 4 stacked PredNet prediction cells used in this paper.}
  \label{fig:PrednetStack}
\end{minipage}
\label{fig:test}
\end{figure}

\citet{lotter2016deep} developed the \textit{Predictive Coding Network} (PredNet) architecture which serves as a baseline for this paper. A single \textit{prediction cell} alongside its entire architecture are presented in \cref{fig:PrednetCell,fig:PrednetStack}, respectively. The single PredNet prediction cell consists of the following elements: an input convolutional layer ($A_l$), a recurrent representation layer ($R_l$), a prediction layer ($\hat{A}_l$), and an error representation ($E_l$). The relationships governing the transitions between elements are shown in the following equations:
\begin{equation}
\label{eqn:prednet}
\begin{split}
A^t_l 
&= 
\begin{cases}
     x_t,\text{  if} \text{ }l=0 \\
     \text{MaxPool}(\text{ReLU}(\text{Conv}(E^t_{l-1}))), \text{ }l>0 \\
\end{cases} \\    
\hat{A}^t_l 
&= 
\text{ReLU}(\text{Conv}(R^t_l)) \\
E^t_l 
&=
[\text{ReLU}(A^t_l-\hat{A}^t_l); \text{ReLU}(\hat{A}^t_l-A^t_l)] \\
R^t_l 
&= 
\text{ConvLSTM}(E^{t-1}_l, \text{Upsample}(R^t_{l+1}))
\end{split}    
\end{equation}

Such prediction cells can be stacked to form larger and more capable recurrent neural networks. In this paper, we consider a 4-layer setup that consists of 4 PredNet prediction cells. More details are presented in \cref{table:prednet}.
\begin{table}[h]
\caption{PredNet Architecture Details.}
\label{table:prednet}
\centering
\begin{tabular}{@{}ll@{}}
\toprule
Hyperparameter & Value\\ \midrule
Number of layers/cells & 4 \\
Number of hidden units per layer & {2, 48, 96, 192} \\
Filter size per layer & {3, 3, 3, 3} \\
Number of parameters & 6 912 766 \\ \bottomrule
\end{tabular}
\end{table}

\subsubsection{Predictive Coding Network with TAAConvLSTM}
In the first proposed approach, we replace the ConvLSTM in the top (4th) PredNet prediction cell with the TAAConvLSTM mechanism. The relationships for the PredNet prediction cell with TAAConvLSTM are as follows:
\begin{equation}
\label{eqn:prednet_taaconvlstm}
\begin{split}
A^t_l 
&= 
\begin{cases}
     x_t, \text{  if} \text{ }l=0 \\
     \text{MaxPool}(\text{ReLU}(\text{Conv}(E^t_{l-1}))), \text{ }l>0 \\
\end{cases} \\ 
\hat{A}^t_l 
&= 
\text{ReLU}(\text{Conv}(R^t_l)) \\
E^t_l 
&=
[\text{ReLU}(A^t_l-\hat{A}^t_l); \text{ReLU}(\hat{A}^t_l-A^t_l)] \\
R^t_l 
&= 
\text{TAAConvLSTM}(E^{t-1}_l, \text{Upsample}(R^t_{l+1}))
\end{split}    
\end{equation}
In our formulation, we define the attention horizon as the $H_a$ most recent past representations. However, we saw an improvement when selecting the $H_a$ past frames uniformly covering a period of \SI{1}{\second}, instead of using the most recent $H_a$ frames covering a period of $H_a \times \Delta_t$ (\SI{0.4}{\second}). This change resulted in a further reduction of moving object vanishing from the predictions. We believe that this choice is more representative of the motion of the agents present in the input scene.
The overview of the hyperparameters for the framework is shown in \cref{table:prednet_taa}.
\begin{table}[h]
\caption{PredNet with TAAConvLSTM Architecture Details.}
\centering
\begin{tabular}{@{}ll@{}}
\toprule
Hyperparameter & Value\\ \midrule
Number of layers/cells & 4 \\
Layers with ConvLSTM & 1, 2, 3 \\
Number of hidden units per layer & 2, 48, 96, 192 \\
Filter size per layer & 3, 3, 3, 3 \\ 
Layers with TAAConvLSTM & 4 \\
Number of heads in TAAConvLSTM & 4 \\
Attention horizon & 4 \\
Attention hidden units $d_k,d_h$ & 0.25 $\times$ hidden units at layer \\
Numbers of parameters & 7 222 094 \\  \bottomrule
\end{tabular}
\label{table:prednet_taa}
\end{table}

\subsubsection{Predictive Coding Network with SAAConvLSTM}
In our second approach, we applied Attention Augmented Convolution introduced by \citet{bello2019attention} to the input-to-state transitions of the ConvLSTM, formulating SAAConvLSTM. The equations for the mechanisms are: 
\begin{equation}
\label{eqn:prednet_saaconvlstm}
\begin{split}
A^t_l 
&= 
\begin{cases}
     x_t, \text{  if} \text{ }l=0 \\
     \text{MaxPool}(\text{ReLU}(\text{Conv}(E^t_{l-1}))), \text{ }l>0 \\
\end{cases} \\ 
\hat{A}^t_l 
&= 
\text{ReLU}(\text{Conv}(R^t_l)) \\
E^t_l 
&=
[\text{ReLU}(A^t_l-\hat{A}^t_l); \text{ReLU}(\hat{A}^t_l-A^t_l)] \\
R^t_l 
&=
\text{SAAConvLSTM}(E^{t-1}_l, \text{Upsample}(R^t_{l+1}))
\end{split}    
\end{equation}
The overview of the hyperparameters for the architecture is shown in \cref{table:prednet_saa}.
\begin{table}[h]
\caption{PredNet with SAAConvLSTM Architecture Details.}
\centering
\begin{tabular}{@{}ll@{}}
\toprule
Hyperparameter & Value\\ \midrule
Number of layers/cells & 4 \\
Layers with ConvLSTM & 1, 2 \\
Number of hidden units per layer & 2, 48, 96, 192 \\
Filter size per layer & 3, 3, 3, 3 \\
Layers with SAAConvLSTM & 3, 4 \\
Number of heads in SAAConvLSTM & 4 \\
Attention horizon & 4 \\
Attention hidden units $d_k,d_h$ & 0.25 $\times$ hidden units at layer \\
Numbers of parameters & 6 705 118  \\ \bottomrule
\end{tabular}
\label{table:prednet_saa}
\end{table}
\subsubsection{PredRNN++}
\begin{figure}[htb]
  \centering
  \centerline{\includegraphics[width=6cm]{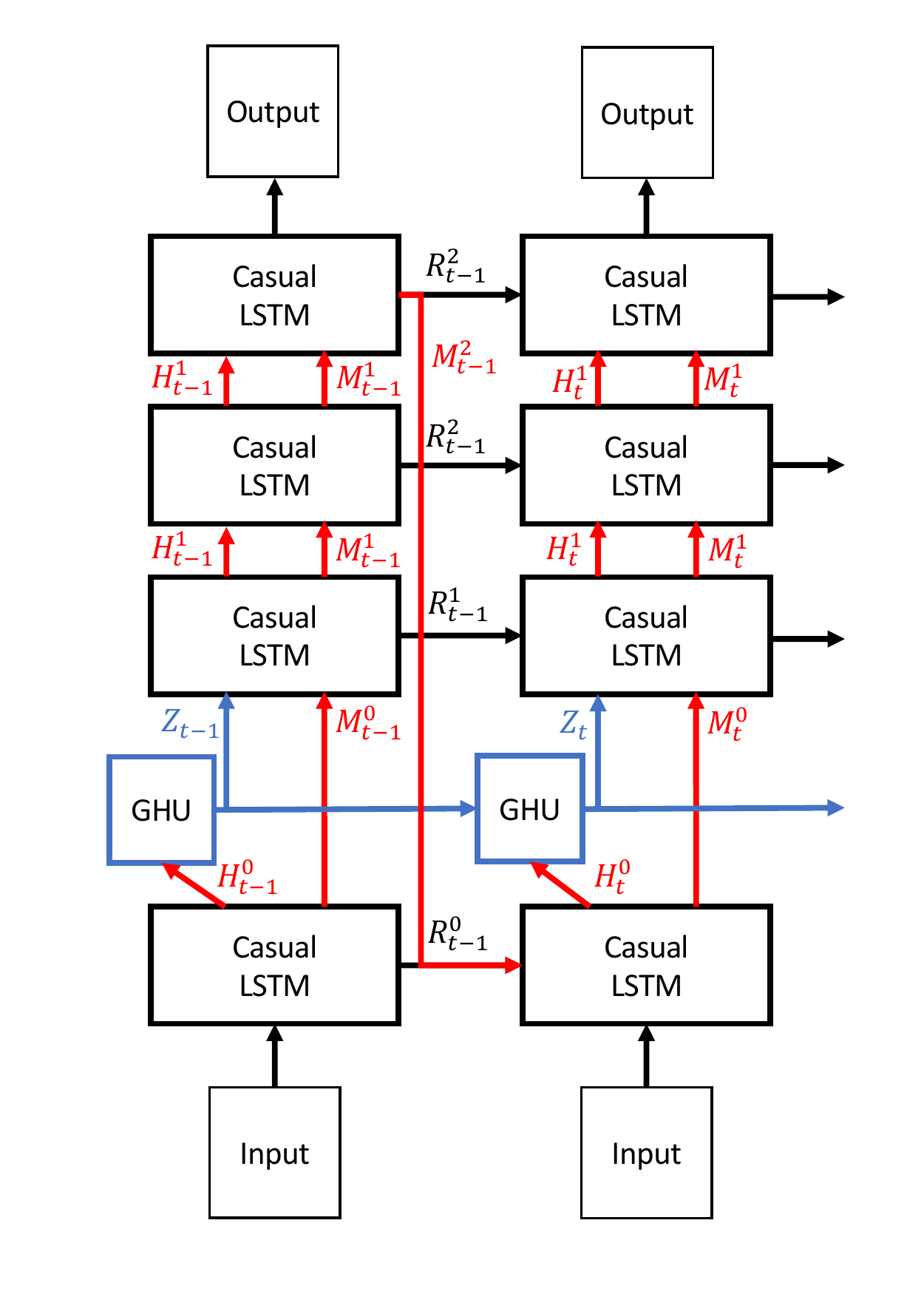}}
  \caption{The PredRNN++ setup used in this paper. Four casual LSTM cells with Gradient Highway Unit (GHU) applied between the first and second layers.}
  \label{fig:PredRNN++}
\end{figure}

\citet{wang2018predrnn++} introduced the PredRNN++ architecture which achieved state-of-the-art results on video prediction tasks. \citet{wang2018predrnn++} developed the casual LSTM, which increases the recurrence depth between time steps and better captures short-term dynamics between frames. Additionally, the architecture includes a Gradient Highway Unit (GHU) to tackle the vanishing gradients issue. The PredRNN++ architecture is shown in \cref{fig:PredRNN++}.
\begin{table}[t!]
\centering
\caption{PredRNN++ Architecture Details.}
\begin{tabular}{@{}ll@{}}
\toprule
\multicolumn{2}{c}{PredRNN++} \\ \midrule
Number of layers/cells & 4 \\
Number of hidden units per layer & 64, 64, 64, 64 \\
Filter size per layer & 5, 5, 5, 5 \\
Stride & 1 \\
Patch Size & 4 \\
Numbers of parameters & 7 249 344 \\ \bottomrule
\end{tabular}
\label{table:predrnn++}
\end{table}

The causal LSTM introduces more nonlinear recurrent transitions and dual memory setup consisting of the temporal memory $\mathcal{C}^k_l$ and spatial memory $\mathcal{M}^k_t$. The relationships governing the casual LSTM are as follows:
\begin{align}
    \nonumber
    \begin{pmatrix}
        g_t \\
        i_t \\
        f_t \\
    \end{pmatrix} 
    &=
    \begin{pmatrix}
        \tanh \\
        \sigma \\
        \sigma \\
    \end{pmatrix}
    W_1 
    *
    [\mathcal{X}_t, \mathcal{H}^k_{t-1},\mathcal{C}^k_{t-1}]\\ \nonumber
    C^k_t 
    &= 
    f_t \circ C^k_{t-1} + i_t \circ g_t \\ 
    \begin{pmatrix}
        g_t'\\
        i_t' \\
        f_t' \\
    \end{pmatrix}
        &=
        \begin{pmatrix}
        \tanh \\
        \sigma \\
        \sigma \\
        \end{pmatrix}
        W_2 * [\mathcal{X}_t, \mathcal{C}^k_{t},\mathcal{M}^{k-1}_{t}] \\ \nonumber
        \mathcal{M}^k_t 
        &= 
        f_t' \circ \tanh (W_3 * M_t^{k-1})+i_t' \circ g'_t \\ \nonumber
        o_t 
        &= 
        \tanh(W_4 * [\mathcal{X}_t, C^k_t, \mathcal{M}^k_t]) \\ \nonumber
        \mathcal{H}^k_t 
        &= 
        o_t \circ \tanh(W_5 * [\mathcal{C}^k_t, \mathcal{M}^k_t] \nonumber
\end{align}
where $W_{1:5}$ are the weights for the convolutional layers.
The GHU relationships are as follows:
\begin{equation}
\centering
\label{eqn:ghu}
    \begin{split}
        \mathcal{P}_t &= \tanh(W_{px} * \mathcal{X}_t + W_{pz} * \mathcal{Z}_{t-1}) \\
        \mathcal{S}_t &= \sigma(W_{sx} * \mathcal{X}_t + W_{sz} * \mathcal{Z}_{t-1} \\
        \mathcal{Z}_t &= \mathcal{S}_t \circ \mathcal{P}_t + (1-\mathcal{S}_t) \circ \mathcal{Z}_{t-1} 
    \end{split}
\end{equation}
where $W_{**}$ are the convolutional layer weights, $\mathcal{S}_t$ is a \textit{Switch Gate}, $\mathcal{P}_t$ is the transformed input, and $\mathcal{Z}_t$ is the hidden state.
The PredRNN++ architecture details used in this paper are presented in \cref{table:predrnn++}.

\subsection{Image Similarity Metric}
\label{sec:is}
The Image Similarity (IS) metric, defined by~\citet{birk2006merging}, determines the picture distance function $\psi$ between two matrices $m_1$ and $m_2$, as follows:
\begin{equation}
\begin{split}
    \psi(m_1,m_2) = \sum_{c \in \mathcal{C}} d(m_1,m_2,c) + d(m_2,m_1,c) 
\end{split}
\end{equation}
where
\begin{equation}
\begin{split}
    d(m_1,m_2,c) = \frac{\sum_{m_1[p]=c} \text{min} \{\text{md}(p_1,p_2)|m_2[p_2]=c\}}{\#_c(m_1)}.
\end{split}
\end{equation}

\noindent $\mathcal{C}$ is a set of discretized values assumed by $m_1$ or $m_2$ which are: occupied, occluded, and free. $m_1[p]$ denotes the value $c$ of map $m_1$ at position $p=(x,y)$. $\text{md}(p_1,p_2)=|x_1 - x_2| + |y_1 - y_2|$ is the Manhattan distance between points $p_1$ and $p_2$.  $\#_c(m_1)= \#\{p_1 \mid m_1[p_1]=c\}$ is the number of cells in $m_1$ with value $c$.